\documentclass[10pt,twocolumn,letterpaper]{article}

\usepackage[table,pdftex,dvipsnames]{xcolor}
\usepackage[framemethod=TikZ]{mdframed}
\usepackage[utf8]{inputenc}
\usepackage{microtype}
\usepackage{graphicx}
\usepackage{subfigure}
\usepackage{multirow}
\usepackage{makecell}
\usepackage{float}
\usepackage{graphbox}
\usepackage{tabularx}
\usepackage[numbers]{natbib}
\usepackage{adjustbox}
\usepackage{authblk}
\usepackage{xcolor,colortbl}
\usepackage{times}
\usepackage[pagebackref=true, breaklinks=true, letterpaper=true, colorlinks,
 citecolor=citecolor, linkcolor=linkcolor, bookmarks=false]{hyperref}
\usepackage{booktabs} 
\usepackage{hyperref}

\usepackage{soul}

\usepackage{amsmath}
\usepackage{amssymb}
\usepackage{mathtools}
\usepackage{amsthm}

\usepackage[capitalize,noabbrev]{cleveref}

\usepackage{stackengine}
\usepackage{svg}
\usepackage{natbib}

\usepackage{titlesec}
\titlespacing*{\section} {0pt}{4ex}{2ex}
\titlespacing*{\subsection} {0pt}{2ex}{2ex}
\titlespacing*{\subsubsection} {0pt}{1ex}{1ex}

\theoremstyle{plain}

\theoremstyle{definition}

\theoremstyle{remark}

\usepackage[textsize=tiny]{todonotes}
\newlength\savewidth

\hypersetup{    colorlinks,
    linkcolor=purple,
    citecolor=gray,
    urlcolor=purple
}

\begin{document}
\vspace{-10em} 
\title{\vspace{-3em}Pinpointing Why Object Recognition Performance\\ Degrades Across Income Levels and Geographies}

\author{Laura Gustafson, Megan Richards, Melissa Hall, Caner Hazirbas, Diane Bouchacourt, Mark Ibrahim}
\affil[]{Fundamental AI Research (FAIR), Meta AI}
\date{}

\maketitle

\begin{abstract}

Despite impressive advances in object-recognition, deep learning systems’ performance degrades significantly across geographies and lower income levels---raising pressing concerns of inequity. Addressing such performance gaps remains a challenge, as little is understood about why performance degrades across incomes or geographies.
We take a step in this direction by annotating images from Dollar Street, a popular benchmark of geographically and economically diverse images, labeling each image with factors such as color, shape, and background. These annotations unlock a new granular view into how objects differ across incomes/regions. We then use these object differences to pinpoint model vulnerabilities across incomes and regions.
We study a range of modern vision models, finding that performance disparities are most associated with differences in \textit{texture, occlusion}, and images with \textit{darker lighting}.
We illustrate how insights from our factor labels can surface mitigations to improve models' performance disparities.
As an example, we show that mitigating a model's vulnerability to texture 
can improve performance on the lower income level.
We release all the \href{https://github.com/facebookresearch/dollarstreet_factors}{factor annotations} along with an \href{https://dollarstreetfactors.metademolab.com/}{interactive dashboard} to facilitate research into more equitable vision systems.

\end{abstract}

\section{Introduction}
\label{sec:introduction}

\begin{figure}[htb]
\centering
\includegraphics[width=0.49\textwidth]{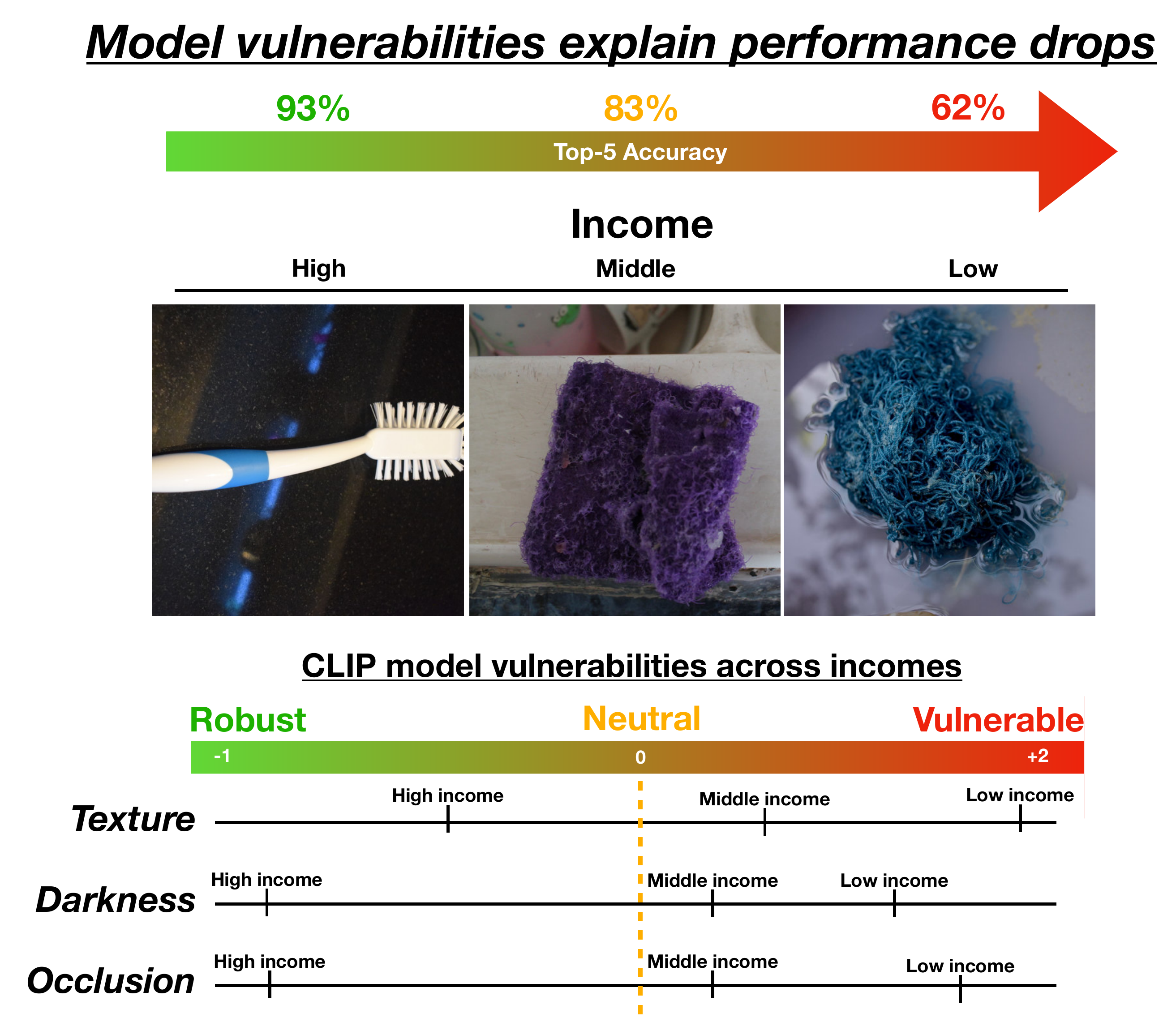}
\caption{CLIP's vulnerability to texture, darker lighting, and occlusion are associated with performance disparities for lower incomes. We rank the most vulnerable factors based on how much more likely a factor is selected among misclassified images than overall. The example images are of \textit{dishbrushes} from Dollar Street.} 
\label{fig:fig1}
\end{figure}

The widespread adoption of object-recognition systems afforded by
advances in deep learning comes with a responsibility: systems should work equally well across groups of individuals.
Previous work demonstrates object-recognition performance
is far from equal across income levels and geographies \citep{de2019does, goyal_fairness_2022, rojas2022the}.
This disparity encompasses publicly available recognition systems, state-of-the-art supervised and self-supervised models. 
Most worrisome among these findings is that the performance degradation disproportionately affects lower income households.
When Artificial Intelligence (AI) systems are deployed in applications such as medical imaging, their biases can lead to disproportional harm. 
For example, models diagnosing COVID-19 were found to rely on geographically-biased features such as the hospital's font to diagnose patients \citep{roberts2021common}.

While existing work measures performance disparities across incomes and geographies, addressing the performance gaps remains a challenge.
Key to progress is understanding \emph{not just that, but why such disparities arise}. 
One hypothesis raised in \citet{de2019does} 
is that objects as well as their environments can vary drastically across regions. 
When factors such as object shape or lighting in a region differ from those commonly seen during training, the shift can cause model performance to drop.
However, no systematic study exists characterizing how such factors vary across regions and incomes. 
Identifying the factors associated with model disparities 
can shed light on research directions to improve performance degradation across incomes and geographies.  

We take a step in this direction by annotating images from Dollar Street \citep{rojas2022the}, a common benchmark for evaluating performance disparities in object recognition systems. 
Dollar Street contains 38k images 
of household objects spanning 54 countries across income levels. 
We annotate each image with factors to mark what makes each distinctive, such as color, pose, shape, and texture.
We first analyze how images vary across incomes and regions using our factor labels in Section \ref{sec:how_object_vary}. We find images of some classes such as \textit{roofs} differ considerably across regions (and incomes) while others (such as \textit{pens}) hardly vary. 

We then investigate how our factor labels can explain model mistakes.
We find an overall correspondence between the distribution of factors per region (and income) and model performance. 
Even for the latest generation of foundation models, such as CLIP \citep{radford_learning_2021},
performance degrades by as much as 25.7\% (top-5 accuracy) across incomes.  
We also compare the performance of other popular models across learning paradigms (self-supervised, supervised), architectures (CNN-,transformer-, MLP-based), as well as large scale pretraining \citep{goyal_vision_2022}. We find remarkably similar vulnerabilities across these popular models.

Next, we precisely rank factors by examining how much more likely they are to appear among misclassifications. A factor much more likely to appear among misclassification suggests a model is vulnerable to the factor. In our analysis we find vulnerabilities in \emph{texture}, \emph{occlusion}, and \emph{darker lighting}
are most associated with models' performance degradation in lower incomes as illustrated in Figure \ref{fig:fig1}. 
We further study class-specific model vulnerabilities, finding strong associations between mistakes for particular classes and factors.
For example, we find that for \emph{sofas}, images labeled with texture are 7.2x more likely to appear among CLIP's mistakes than overall, suggesting texture bias is a vulnerability.

Finally, we study whether we can use robustness techniques to make fairness improvements. We show that mitigating the texture vulnerability surfaced by our analysis can improve performance disparities across incomes/regions in Section \ref{sec:improving_disparity_texture}. We find a model trained to mitigate texture bias not only performs better overall +0.8\% (top-5 accuracy), but on the relevant subset of images (those marked with \texttt{texture}), improves accuracy by 4.1\% for \texttt{low} incomes. This suggests factor labels not only explain mistakes, but can even reveal promising mitigations to combat the 
disparities we observe in vision models today.
Along with our analysis, we release all the factor annotations 
with an interactive dashboard to enable research facilitating more responsible, equitable vision system.

\begin{figure}[htb]
\centering
\includegraphics[width=0.49\textwidth]{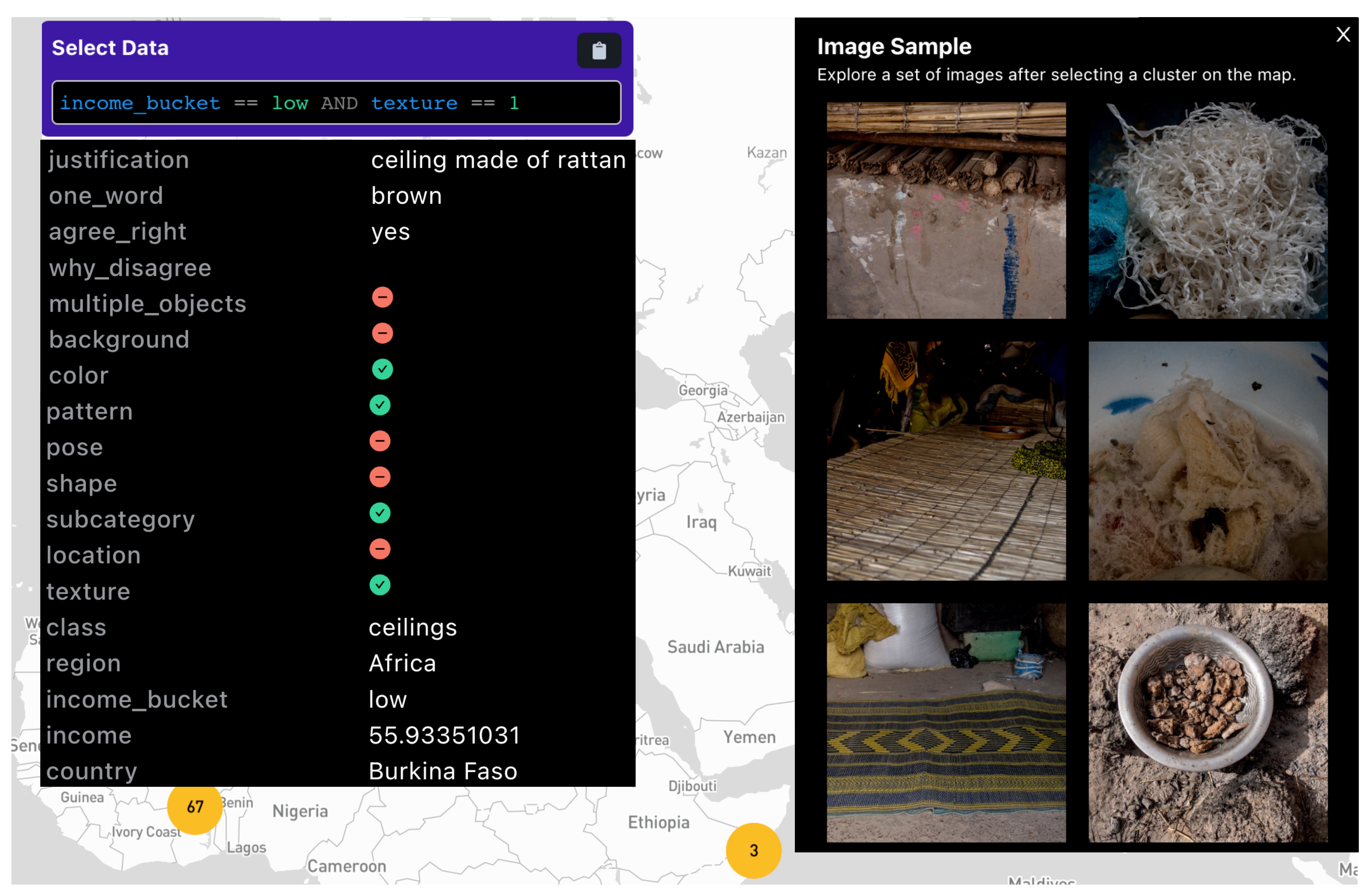}
\caption{A screenshot of our \href{https://dollarstreetfactors.metademolab.com/}{interactive dashboard} allowing researchers to query subsets, visualize images, and view our corresponding factor labels.} 
\label{fig:fig-dashboard}
\end{figure}

To summarize, our contributions are 1) we annotate all of Dollar Street images with distinctive factor labels such as pose, background, and color, 2) we explain performance disparities in models (including CLIP) using our factor annotations to reveal vulnerabilities in texture, occlusion, and darker lighting, 3) we demonstrate mitigating the vulnerability to texture can improve performance disparities across incomes and geographies, 4) we release all our factor annotations with a dashboard (Figure \ref{fig:fig-dashboard}) allowing researchers to interactively query and visualize image factor labels to spur research into equitable vision systems.

\section{Annotating Dollar Street with\\ factor labels}
\vspace{-0.5em}
The Dollar Street dataset is a common computer vision benchmark for classifying everyday objects (e.g \textit{armchairs, toothbrush, pens}) across incomes and geographies. Households across the world upload images of the specified objects. These images are labeled with the object class, location, and income of the household. 
The income is standardized on an international scale by DollarStreet \citep{DSIncome}.
We use the procedure described in \citep{goyal_fairness_2022} to aggregate household incomes into buckets: \texttt{high}, \texttt{medium} and \texttt{low}, and group countries into regions: \texttt{Asia}, \texttt{Africa}, \texttt{Europe}, \texttt{The Americas}. 
Table \ref{app:image_region_combination_number} in Appendix \ref{app:ds_stats} shows the number of images per income bucket/region pair.
\vspace{-0.25em}
\subsection{Annotation Procedure}
\vspace{-0.25em}
In order to explain the degradation in model performance across incomes or geographies, annotators labeled images in Dollar Street with the factors distinguishing each image.
We select all 14k images overlapping with classes in the ImageNet-21k taxonomy from \citet{ridnik2021imagenet}, using the mapping from DollarStreet classes to ImageNet synsets from \citet{goyal_fairness_2022}.
We follow the same annotation procedure as in \citet{idrissi2022imagenet}.
Since it's challenging to accurately label an image in isolation, we ask annotators to label how each image differs from a fixed set of three prototypical images chosen for each class. We define prototypical images for each class as those correctly classified by a ResNet-50 model with the highest confidence. We curate a list of sixteen potential factors that can distinguish an image from the prototypical images for its class. These factors include pose, various forms of occlusion, size, style, type or breed. A full list is shown in Figure \ref{fig:factors_income_region}. Annotators select any number of factors they believe best distinguish each image.
In addition, we ask annotators to provide text descriptions to account for factors outside the sixteen factors
we provide, and ask if they agree with the original class label (see \ref{sec:affordances} for further discussion of how this was used). A more detailed description of the annotation setup and prototypical images is in Appendix \ref{app:annotation_all}. 
\vspace{-0.225em}
\subsection{Factor label statistics}

\begin{figure}[htb]
\centering
\adjustbox{trim=0.3cm 0.5cm 1.5cm 0.5cm}{
\includesvg[width=0.55\textwidth,pretex=\fontsize{8pt}{8pt}]{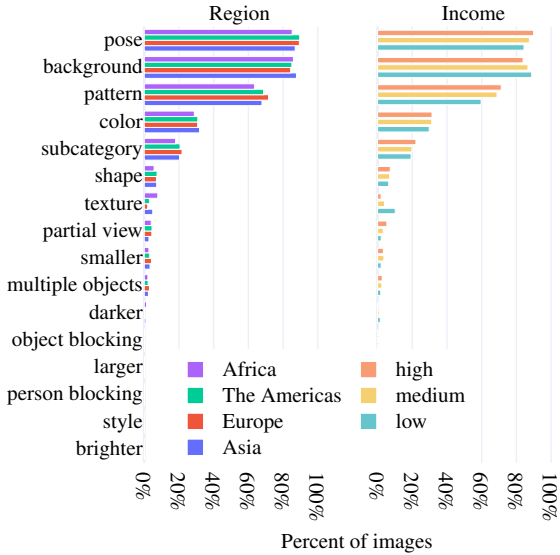}
}
\caption{\textbf{Pose, background, and pattern are the most commonly selected factors}. Figure shows the percent of images by region and income that were labelled with each factor. Annotators labelled each image with the factors that most distinguished each image from the prototypical images of its class.} 
\label{fig:factors_income_region} 
\end{figure}

We first explore how frequently each factor was selected across income levels and regions.
In Figure \ref{fig:factors_income_region}, we plot the distribution of factor labels across regions and income buckets. On average, annotators chose 3.2 factors per image (standard deviation 1.2). The two most correlated factors are \textit{color} and \textit{pattern}, with a correlation coefficient of 0.19.

\paragraph{Across all incomes, \textit{pose}, \textit{background}, and \textit{pattern} are the most selected factors.}
For most factors, there is only a minor difference in the frequency that a factor was selected across income buckets.
For \textit{texture}, however, there's a noteworthy difference across income levels with 10.2\% of images in the \texttt{low} income bucket labeled with \emph{texture} compared to only 4.2\% for  \texttt{medium} and 2.0\% for \texttt{high} income buckets.
This implies \textit{texture} is 5x
more likely to 
be selected within the \texttt{low} income bucket (relative to the \texttt{high} income bucket), a stark difference. 

\paragraph{The most commonly selected factors are consistent across regions and incomes.}
 Similar to our observation across income buckets, the most striking difference across regions is for \textit{texture}. \textit{Texture} is selected for 7.8\% of images in  \texttt{Africa} but for only 2\% of images in  \texttt{Europe}---a 4x difference. The similarity of emerging patterns in factors across incomes to those across regions suggests that income bucket variation differences are also exhibited across geographies. This can in part be explained by the relative rates of co-occurrence of regions and income buckets in DollarStreet, see Table \ref{app:image_region_combination_number} in Appendix \ref{app:ds_stats}.

\subsection{Controlling for regional differences in raters' perceptions}
\label{sec:affordances}

Challenges naturally arise when running such a large annotation procedure. In our case, there can exist regional perceptions of the semantic meaning of every object label. Indeed, the Dollar Street object class labels were originally collected from the household members who took the image, rather than assigned retroactively, which means that regional perceptions could be a source of variance in the dataset. To control for the effect of such perceptual differences across regions, we created a second annotation task. 
To select the images for the task, we analyzed the results of the original factor annotations, which included an option for annotators to disagree with the object label. We selected the subset of images where the annotator disagreed with the label, which totaled 2,476 images, or 18.1\% of the Dollar Street dataset. For the second task, we collected additional annotations for this disputed subset, asking annotators sourced from 6 countries (5 continents) whether they agreed or disagreed with the object label. In total, there were 79 annotators who provided 12,507 label annotations. See Appendix \ref{app:label_agreement_overveiw} for examples of annotator disagreements, and classes with the highest and lowest levels of label disagreement. 

We then compared the levels of disagreement between annotators from the region where the image was taken (source region) and annotators who were from other regions. If there were region-specific biases in the label, we would expect a much higher rate of disagreement for annotators not from the source region. For images in the second task, we found annotators from both the source and other regions disagreed with the original label at similar rates (an average of 49.1\% and 46.8\% respectively). 
This consistency suggests regional differences in label perceptions do not constitute a significant source of class variation in Dollar Street. As a result, we can turn to our factor annotations to characterize how class images vary across regions and incomes. 

\section{\raggedright How do objects vary across incomes and geographies?}
\label{sec:how_object_vary}

Leveraging the factors gathered in our annotation process, we can quantitatively assess how classes differ across incomes and regions. For example, we can identify classes that change the most across incomes (e.g. ceiling), as well as the corresponding factors that best differentiated \texttt{high} or \texttt{low} income ceilings (pose, subcategory, and texture). 

To compare how classes differ across two regions (or incomes), we compute the distribution of factors for each class. Specifically, for every pair of regions (or incomes) we normalize the distributions per class then measure the Jensen-Shannnon Distance (JSD) to characterize how images differ across regions (or incomes). Note the Jensen-Shannon Distance is the square root of Jensen-Shannon divergence between two distributions, and is a standard metric to compare discrete distributions \citep{jsd2003}. A large JSD distance between two regions for a class indicates images differ across those regions.

\begin{table}
    \centering
\begin{tabularx}{.49\textwidth}{lll}
\toprule
               class & income bucket &              differentiating factors \\
\midrule
\multicolumn{3}{c}{Most stark difference by \textit{income}} \\
\midrule
               \small{roofs} &    \small{\texttt{low}} \tiny{vs} \small{\texttt{high}}&       \small{subcategory, pose, smaller} \\
            \small{ceilings} &   \small{\texttt{low}} \tiny{vs} \small{\texttt{high}}&       \small{pose, subcategory, texture} \\
             \small{diapers} &   \small{\texttt{low}} \tiny{vs} \small{\texttt{high}} &            \small{color, shape, texture} \\
              \small{radios} &   \small{\texttt{low}} \tiny{vs} \small{\texttt{high}} &        \small{color, shape, subcategory} \\
              \small{floors} &   \small{\texttt{low}} \tiny{vs} \small{\texttt{high}}&           \small{texture, pose, pattern} \\
\midrule
 \multicolumn{3}{c}{Most stark difference by \textit{region}} \\
\midrule
\small{chickens} &          \texttt{\scriptsize{Asia}} \tiny{vs}  \texttt{\scriptsize{Europe}} &         \small{partial view, color, shape} \\
 \small{chickens} &        \texttt{\scriptsize{Europe}} \tiny{vs}  \texttt{\scriptsize{Africa}} &          \small{pose, partial view, color} \\
  \small{diapers} & \texttt{\scriptsize{Americas}} \tiny{vs} \texttt{\scriptsize{Africa}} &          \small{pose, color, partial view} \\
\small{pet foods} &    \texttt{\scriptsize{Asia}} \tiny{vs} \texttt{\scriptsize{Americas}} &            \small{color, texture, pattern} \\
\small{pet foods} &          \texttt{\scriptsize{Asia}} \tiny{vs}  \texttt{\scriptsize{Europe}} &        \small{pattern, subcategory, color} \\
            
\bottomrule
\end{tabularx}
    \caption{Classes with most stark differences in factor distributions by Jensen-Shannon Distance (JSD) across incomes/regions.}
    \label{tab:jsd_factor_diff}
\end{table}

\paragraph{Some classes' factors vary significantly across incomes and regions; others remain consistent.} 

\begin{figure}
    \centering
    \includegraphics[width=0.45\textwidth]{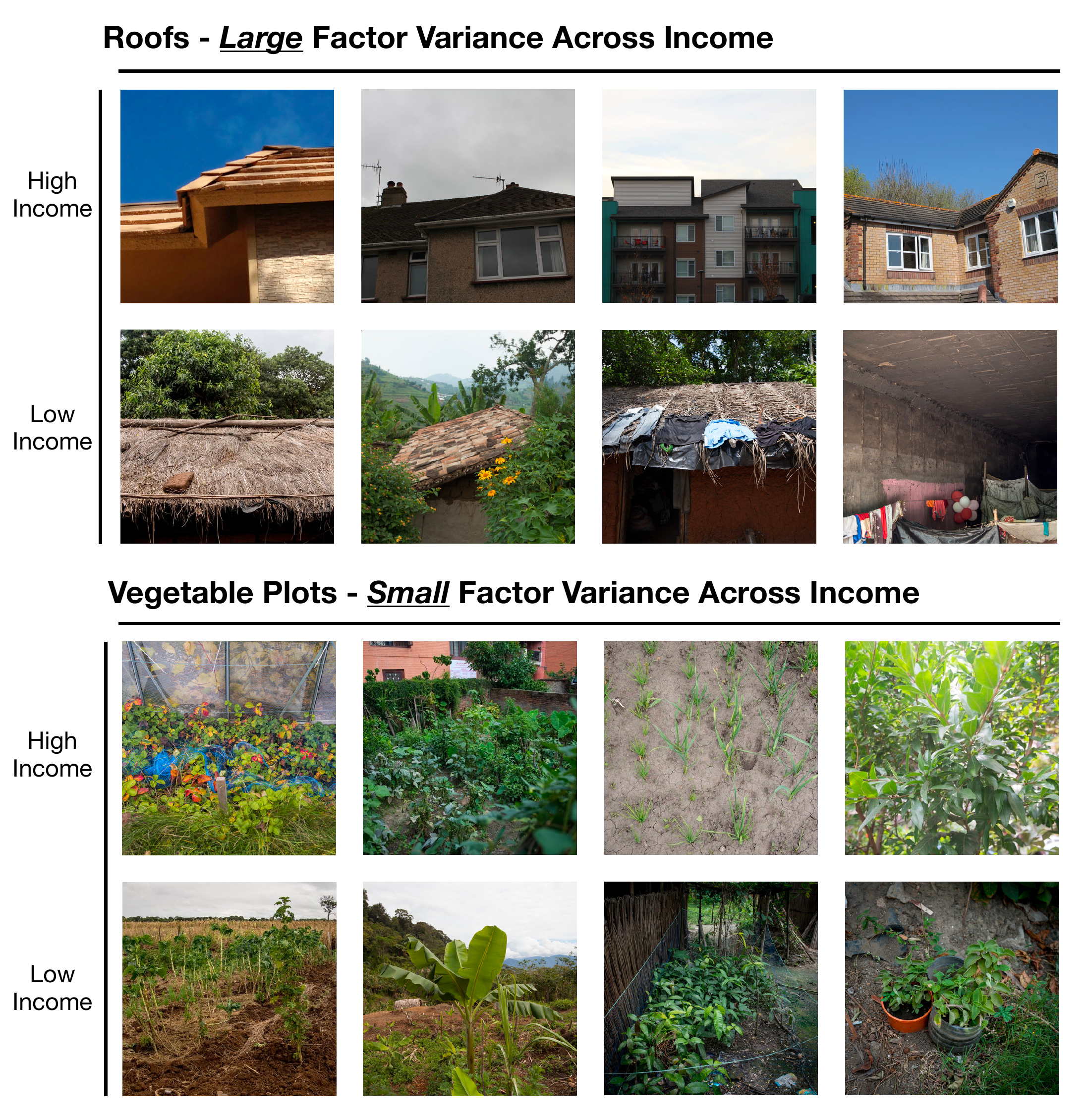}
    \caption{Example images for roofs, the class with the largest factor variation across incomes, and vegetable plots, the class with the smallest factor variation across income.}
    \label{fig:image_examples_factor_income}
\end{figure}

In Table \ref{tab:jsd_factor_diff}, we show the most starkly different classes across incomes/regions, along with their most distinguishing factors. Consistently, we find the largest differences are between \texttt{low} and \texttt{high} incomes, but don't find a consistent pattern with regions. For classes with the largest differences across regions/incomes, we find these differences to be significant with an average JSD more than twice the median JSD of all classes (across incomes/regions). Among classes differing most across regions are those relating to 
animals (\emph{chickens, pet foods}); those differing most across incomes relate to building structures (\emph{roofs, ceiling, floors}). We report full tables for the most similar and dissimilar classes across regions and incomes in Appendix \ref{app:object_variation}. In contrast, other classes don't vary across incomes/regions such as \emph{vegetable pots}, \emph{phones}, and \emph{pens} as shown in Figure \ref{fig:image_examples_factor_income}. This suggests that while some classes can vary drastically across incomes or regions, others are quite similar.

\section{Modern models' performance degrades across incomes and geographies}
\label{sec:model_performance_degrades}

We first study the popular foundation model CLIP, which has been shown to have
strong zero-shot performance on several classification benchmarks \citep{radford_learning_2021}. 
CLIP is trained on 400M text-image pairs using a text encoder and an image encoder enabling a user to perform zero shot classification for any image. 
Here we prompt the model using the set of Dollar Street classes for each image to generate predictions. Our evaluation setup is described in detail in Appendix \ref{app:eval_setup}.

\begin{figure}
    \centering
    \includegraphics[height=1.5in]{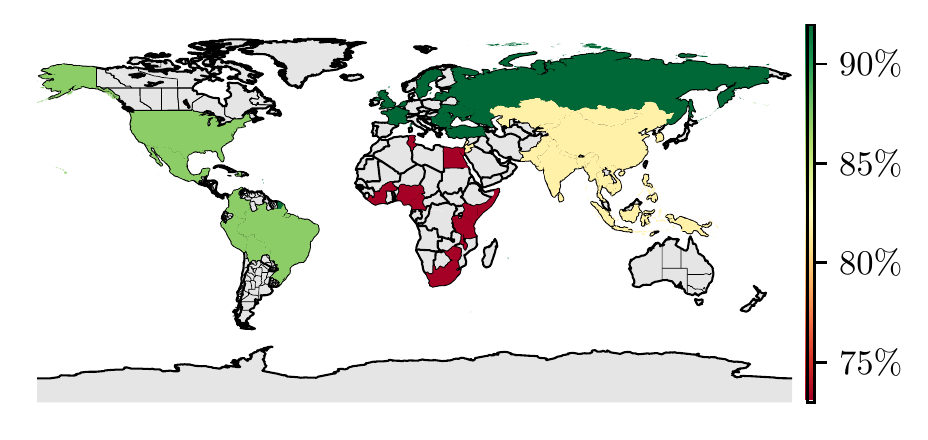}
    \\ 
 \begin{tabular}{c | c }

 Income &  Top 5 \\
 Bucket &  Accuracy \\
\hline \hline 
  Low &  66.9  \\
\hline
 Middle &  83.4  \\
\hline
 High &  92.6 \\
\end{tabular}  
\\ 
 
\caption{\textbf{CLIP performance degrades across incomes and geographies}. Shows CLIP ViT B/32 top-5 accuracy per region and income bucket. Color indicates model performance per region.} 
\label{fig:acc_region_income} 
\end{figure}

We show the performance of CLIP ViT B/32 on DollarStreet across income and geographies in Figure \ref{fig:acc_region_income}.
Despite impressive performance across several other classification and robustness benchmarks, we find performance drops by 25.7\% from \texttt{high} to \texttt{low} incomes. We observe similar drops across regions. Overall the performance disparities for CLIP are comparable to those observed for other models in prior work \citep{goyal_fairness_2022}, suggesting 
performance disparities remain even in the latest generation of foundation models.
Next we conduct a more comprehensive analysis of performance disparities across other model types.

\newsavebox\mybox
\newcommand\Includegraphics[2][]{\sbox{\mybox}{  \includegraphics[#1]{#2}}\abovebaseline[-.5\ht\mybox]{  \addstackgap{\usebox{\mybox}}}}

\begin{figure}[htb]
\centering
\includesvg[width=0.49\textwidth,pretex=\fontsize{8pt}{8pt}]{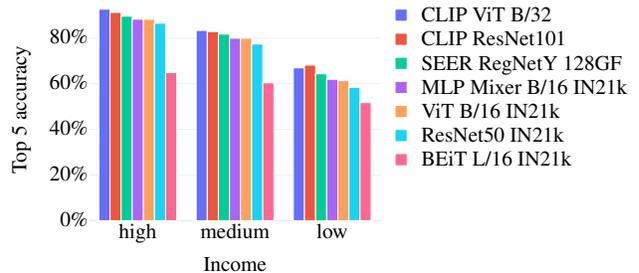}
    \caption{\textbf{Across architectures and learning procedures, model performance degrades similarly for lower incomes}. Bars indicate top-5 accuracy.}
    \label{fig:inc_all_models}
\vspace{-1em}
\end{figure}

\begin{figure}[htb]
\centering
\includesvg[width=0.49\textwidth,pretex=\fontsize{8pt}{8pt}]{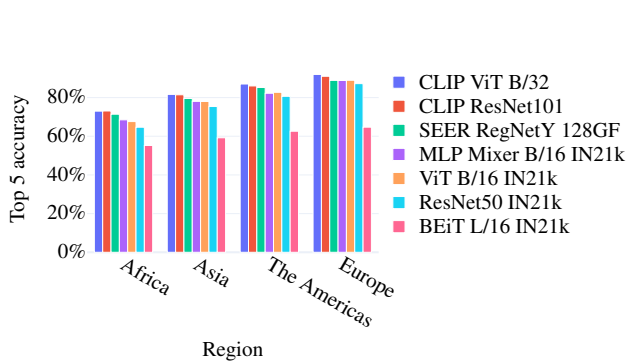}
    \caption{\textbf{Across architectures and learning procedures, model performance degrades similarly across geographies}. Bars indicate top-5 accuracy.}
    \label{fig:region_all_models}
\end{figure}

\paragraph{Performance inequities are pervasive across architectures and training methods.}
We expand our study to encompass models across architectures (convolutional, transformer, and feedforward), learning paradigms (self-supervised and supervised), and pretraining datasets of various sizes (up to 1 billion images).
To generate predictions on Dollar Street, the images are pretrained or finetuned on ImageNet21k and
we use the same mapping from \citet{goyal_fairness_2022} and detail our evaluation procedure in Appendix \ref{app:eval_setup}.
We compare the performance of the set of models across incomes and regions in Figures \ref{fig:inc_all_models} and \ref{fig:region_all_models}.
We find most models have comparable drops in accuracy across incomes and regions despite model differences. BEiT performs worse overall, but still exhibits similar trends in performance gaps across incomes and geographies \citep{bao2021beit}. This suggests even large scale pretraining in models such as SEER \citep{goyal2021self}, modern architectures such as ViT \citep{ridnik2021imagenet} or MLP Mixer \citep{tolstikhin2021mlp}, and self-supervised learning still don't address performance inequities.
Why do such consistent disparities arise? Next we study whether factors labels can explain the performance disparities in modern vision models.

\section{Explaining model performance disparities with factor labels}
\label{sec:explain_model_mistakes}

We now study how our factor labels can explain the model performance disparities we observed across regions/incomes. 
After ruling out variables such as image quality and class imbalance in training, we demonstrate how our factor labels can surface specific model vulnerabilities associated
with degradation in performance across regions/incomes.

\subsection{Controlling for factors not captured in our annotations} 
\label{sec:controls}
\paragraph{Performance disparities are not explained by image quality or training data class imbalance.} As image quality has been shown to impact the performance of facial recognition models \citep{xu2014low}, we first investigate whether image resolutions differ across regions and effect model performance. \citet{rojas2022the} found very minor differences in average image quality across region in DollarStreet. We take this a step further and find no strong correlation ($<0.05$ Pearson’s correlation coefficient) between image DPI and model performance (top-5 accuracy).
Next, since class imbalance in training can skew model performance, we also investigate the extent to which class imbalance affects the disparities we observe.
For ImageNet-21K pretrained or finetuned models (ViT, ResNet, MLPMixer, BEiT, and SEER), we calculate the Pearson correlation between number of images in each class for ImageNet-21K and model's top-5 accuracy. 
We found similarly weak correlations for all models, with coefficients less than 0.25 for top-5 accuracy (all values reported in Appendix \ref{app:explain_model_mistakes}). These results suggest that variation in image quality and pretraining class imbalance explain very little of the variation model mistakes in Dollar Street. 
Next, we examine whether our factor labels can explain model performance disparities.

\subsection{Variation in factor labels are indicative of performance disparities}

To assess whether our factor labels are indicative of model performance disparities, we measure whether larger differences in factor labels across incomes/regions correspond to larger degradations in model performance. 
Specifically,
for each class we measure the Jensen Shannon Distances
(JSD) between the factor label distributions of every pair of
income buckets (and regions).
This quantifies how images in a classes vary across income bucket (or region) pairs 
according to our factor labels. 

\begin{figure} 
\centering
\includesvg[width=0.45\textwidth,pretex=\fontsize{8pt}{8pt}]{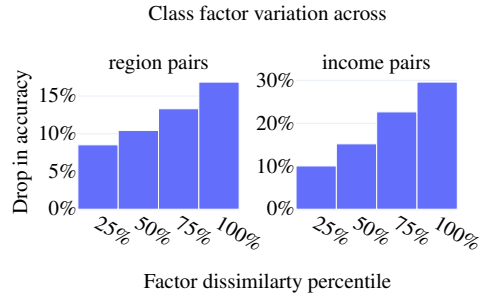}

\caption{\textbf{
Differences across regions/incomes measured by our factor labels are indicative of performance disparities.}
The factor dissimilarity measures the distance (JSD) in factor label distributions across two regions/incomes for a specific class. The drop in accuracy measures the drop in performance for the given class across two regions/incomes.}

\label{fig:jsd_quartiles} 
\end{figure}

Next, we calculate whether larger differences in images across incomes/regions correspond to larger disparities in model performance.
We find as a class varies more across income pairs (according to the JSD of factor distributions), model performance gaps also increase, as shown in Figure \ref{fig:jsd_quartiles}. 
For example, classes that differ most across incomes (top quarter) suffer a 3x drop in accuracy compared to classes that differ the least across incomes (bottom quarter).
We find a similar but less stark trend across regions shown in Figure \ref{fig:jsd_quartiles}.
These results suggest \textit{our factor labels can explain model performance disparities across regions/incomes}.

\subsection{Model performance disparities are most associated with texture, darker lighting, and occlusion}

To more precisely assess which factors are most associated with mistakes, 
we use the same error ratio metric from \citet{idrissi2022imagenet} to measure the association between each factor and model errors. 
Specifically, the error ratio for a factor quantifies how much more or less likely a factor is to appear among a model's misclassified samples as 

\begin{equation}
   \frac{P(\text{factor X } | \text{ model errors}) - P(\text{factor X})}{P(\text{factor X})}
\end{equation}

An error ratio greater than zero indicates how much more likely a factor is to appear among misclassified samples suggesting the factor is associated with model mistakes. For example, an error ratio of 2x indicates a factor is 2x more likely to be selected among misclassified samples than overall.
An error ratio less than zero indicates a factor is less likely to appear among misclassified samples suggesting the model is robust to the factor.
Since some factors are selected only for a few images, we exclude factors selected for five or fewer images in our analysis. 
Doing so excludes style and brightness.

\begin{figure}[htb]
\vspace{-1em}
    \centering
    \includegraphics[width=0.49\textwidth,trim={0 0 0 1cm},clip]{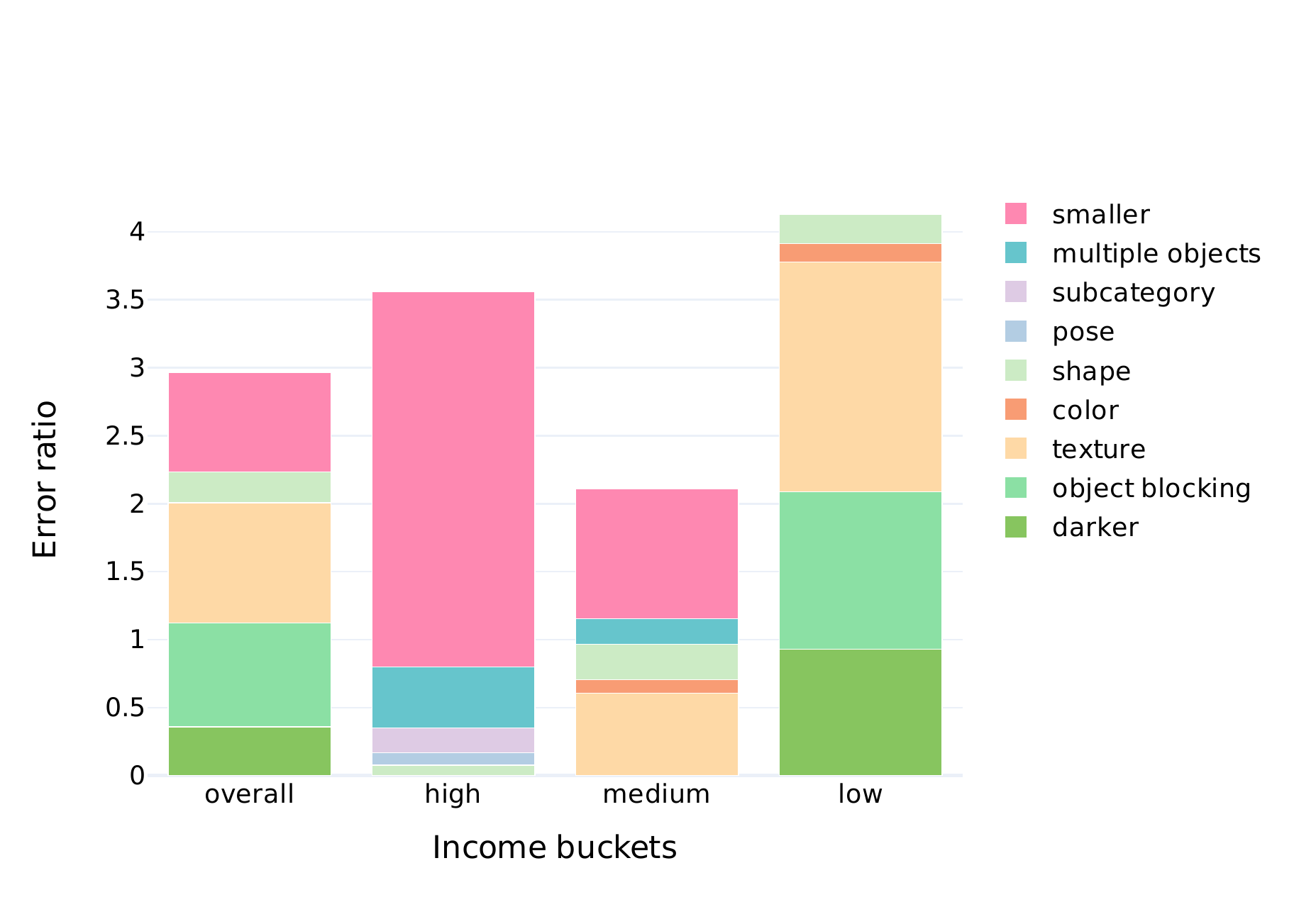}
    \caption{\textbf{CLIP is most vulnerable to texture, occlusion, and darker lighting for low incomes.} Figure shows the five factors most associated with model mistakes across incomes. The area within each bar represents the error ratio measuring how much more likely a factor is selected among the model's misclassified samples.}
    \label{fig:top_5_factors_per_income_bucket}
\end{figure}

\label{par:texture_vulnerability}
\paragraph{Texture, occlusion, and darker lighting are most associated with model disparities across incomes and regions.}
We examine the five factors most associated with model mistakes (measured using error ratio).  
Overall, we find mistakes for CLIP with a ViT-B/32 encoder are most associated with \textit{texture}, \textit{occlusion}, or objects appearing \textit{too small} as shown in Figure \ref{fig:top_5_factors_per_income_bucket}.
For example, \textit{texture} appears +0.88x more among CLIP's mistakes than overall. Similarly, occlusion appears +0.76x and smaller +0.73x more so among mistakes. We conduct $\chi^2$-test to verify such differences in factor prevalence are statistically significant (see Appendix \ref{app:explain_model_mistakes}).

\begin{figure}[htb]
    \centering
    \includegraphics[width=0.49\textwidth,trim={0 0 0 1cm},clip]{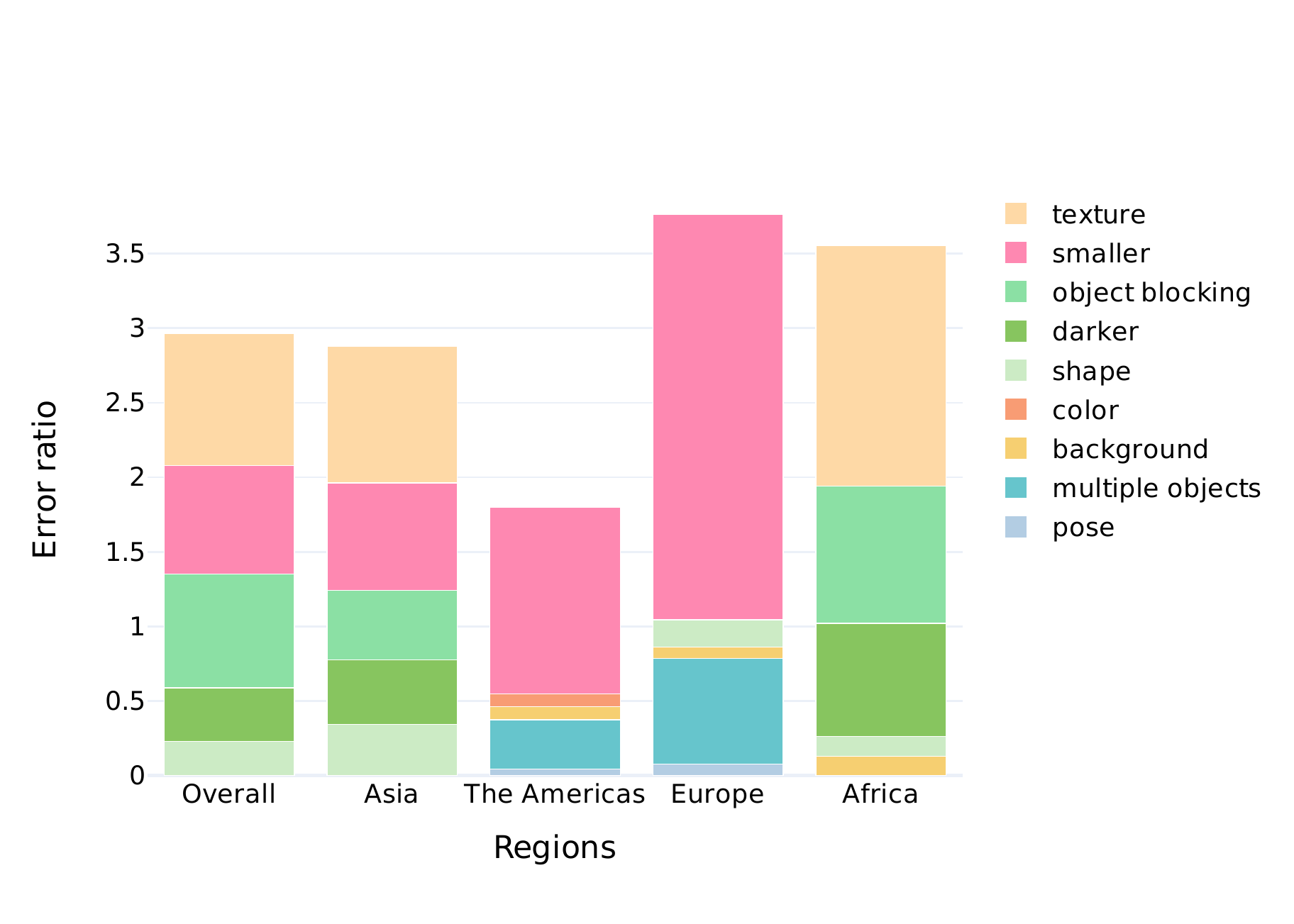}
    \caption{\textbf{CLIP is vulnerable to texture and occlusion in Africa and Asia.} The five factors most associated with model mistakes in each region.  The area within each bar represents the error ratio measuring how much more likely a factor is selected among the model's misclassified samples.}
    \label{fig:top_5_factors_per_region}
\vspace{0.5em}
\end{figure}

We find these vulnerabilities also explain the performance
disparities we observe for \texttt{low} incomes and regions with lower performance (\texttt{Africa}).
In Figure \ref{fig:top_5_factors_per_income_bucket} we show the five factors most associated with model mistakes across incomes (and across regions in Figure \ref{fig:top_5_factors_per_region}). 
We find in the \texttt{low} income bucket,
texture has the largest error ratio with \textit{texture} 1.7x more likely to be selected among misclassifications in the \texttt{low} income bucket.
On the other hand, for the \texttt{high} income bucket misclassifications are associated with quite different factors. For example, smaller objects are most associated with mistakes in the \texttt{high} income bucket. 
We find a similar trend for across regions with \textit{texture}, \textit{occlusion}, and \text{darker lighting} most associated with mistakes in the region of \texttt{Africa}.
We also detail model strengths by measuring factors that are much less likely to appear among mistakes in Appendix \ref{app:explain_model_mistakes}. For example, we find CLIP is much less likely to misclassify images with \textit{partial views} of objects.

\subsubsection{ Some model vulnerabilities are class-specific}
\label{sec:explain_specific_class_mistakes}
Beyond explaining performance disparities across regions or income buckets overall, our factor labels enable us to explain vulnerabilities
for specific classes.
We show in Table \ref{tab:class_vulnerabilities} the factors most associated with mistakes for classes with the lowest performance across regions.
We find the lowest performance is for images in the region of \texttt{Africa} with each class exhibiting 
class-specific vulnerabilities. 
For \textit{shaving}, \textit{shape} is most associated with the low performance as it's 5.8x more likely to appear among mistakes in \texttt{Africa}.
For \textit{sofas}, \textit{texture} is 7.2x more likely to appear among mistakes.
We find a similar pattern for classes with the low performance across incomes in Appendix \ref{app:explain_model_mistakes} and \ref{app_sec:worst_classes}.
These strong shifts in error ratios point to class-specific vulnerabilities. 

Next we study vulnerabilities across a range of model architectures and learning procedures.

\begin{table}
\centering
\begin{adjustbox}{width=0.49\textwidth}
\begin{tabular}{lll}
\toprule
        Class & Region &                            Factors most associated with mistakes \\
\midrule
      shaving &  \texttt{Africa} &          shape (+5.8x), pattern (+0.2x), background (+0.1x) \\
        sofas &  \texttt{Africa} &              texture (+7.2x), pattern (+0.3x), pose (+0.1x) \\
    bathrooms &  \texttt{Africa} &             background (+1.1x), pose (+0.2x), color (+0.1x) \\
kitchen sinks &  \texttt{Africa} &             color (+0.6x), background (+0.2x), pose (+0.2x) \\
      showers &  \texttt{Africa} & background (+1.2x), pose (+0.4x), multiple objects (-1.0x) \\
\bottomrule
\end{tabular}
\end{adjustbox}
\caption{Class-specific vulnerabilities surfaced by our factor labels. We show vulnerabilities for the classes with lowest regional performance.}
\label{tab:class_vulnerabilities}
\end{table}

\section{\raggedright The effect of architecture and training procedure on mistake types}

The factor labels also allow us to compare vulnerabilities across models. We first study vulnerabilities across a range of vision models then we show mitigating the most common vulnerability (to \textit{texture}) can improve performance disparities across incomes and geographies.

\subsection{Comparing vulnerabilities across architectures and training procedures}
\begin{figure}[htb]
\centering
        \includegraphics[width=0.49\textwidth,trim={0 0 0 2cm},clip]{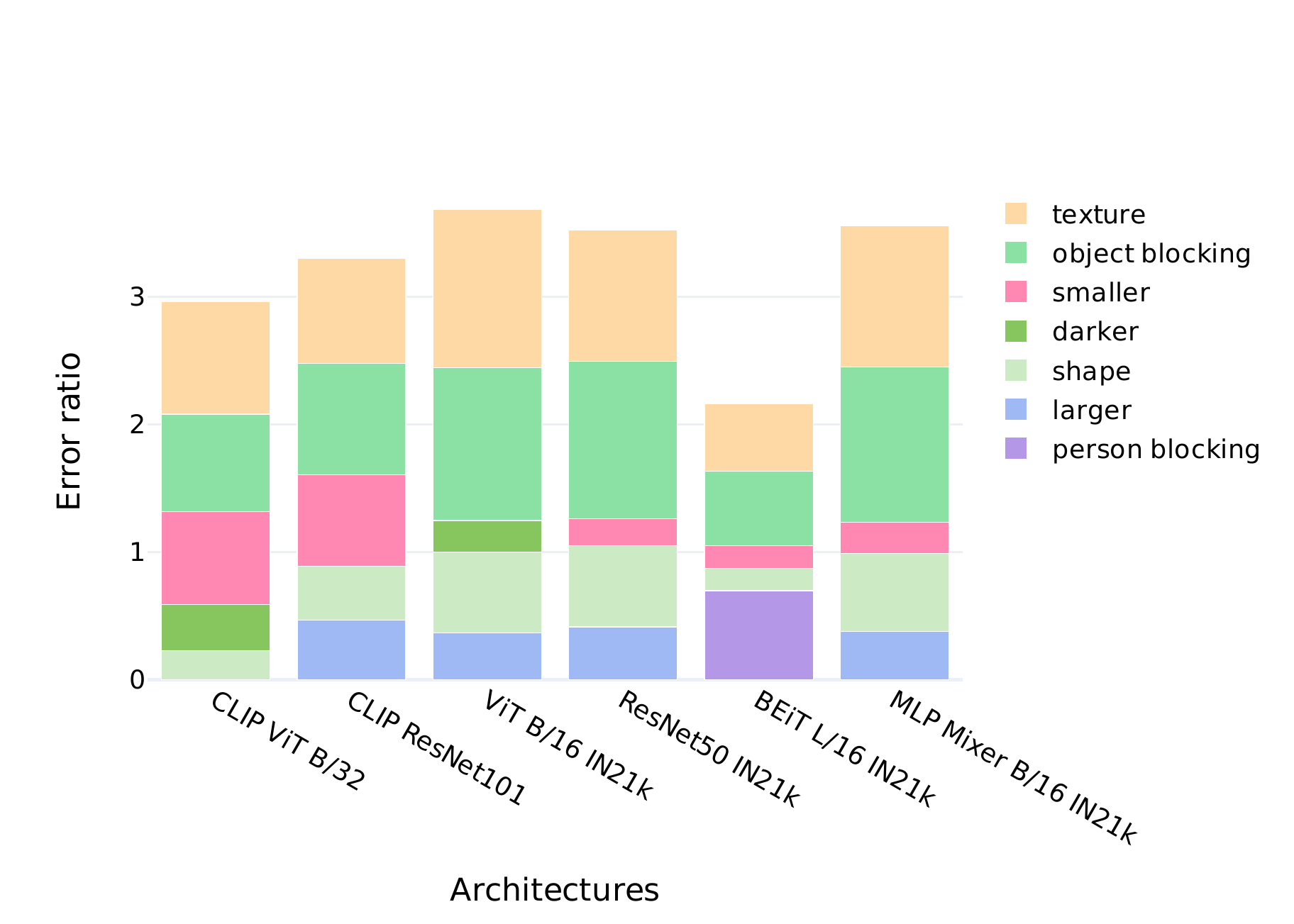}
    \caption{\textbf{Model vulnerabilities are similar across architectures and learning procedures.} Figure shows the five factors that are most associated with each model's mistakes where the area of each bar indicates the error ratio.}
    \label{fig:top5_factors_per_model}
\end{figure}
In Figure \ref{fig:top5_factors_per_model}, we examine model vulnerabilities across architectures and training methodologies. We find that, for all models, \emph{texture}, \emph{occlusion}, and \emph{shape} are consistently among the factors most associated with model mistakes. While texture is known to be a bias specifically for convolutional-models \citep{geirhos_imagenet-trained_2022, hermann2020origins}, we find regardless of architecture or training procedure models have similar vulnerabilities.

\subsection{Improving performance disparities by mitigating texture bias}
\label{sec:improving_disparity_texture}

\begin{table}[]
\centering
\begin{tabular}{@{}lllll@{}}
\toprule
\multirow{2}{*}{Top-5 Accuracy} & \multirow{2}{*}{Overall} & \multicolumn{3}{l}{\textit{\small{Images Marked with Texture}}} \\ \cmidrule(l){3-5} 
                                &                          & \small{low income}        & \small{Africa}           & \small{Asia}             \\ \midrule
ResNet-50                       & 32.2                     & 25.2              & 18.6             & 24.4             \\
Texture debiased                & \textbf{33.0}            & \textbf{29.3}     & \textbf{21.6}    & \textbf{25.0}    \\ \bottomrule
\end{tabular}
\caption{Texture debiasing \citep{geirhos_imagenet-trained_2022} can improve performance across low income buckets and regions with lower performance for images marked with texture as a distinctive factor.}
\label{tab:texture_debiasing}
\end{table}

Our analysis in Section \ref{par:texture_vulnerability} reveals \textit{texture} is most associated with model's performance discrepancy across incomes and geographies. Can we improve this performance disparity by mitigating models' reliance on texture? To assess this, we compare in Table \ref{tab:texture_debiasing} the performance of a standard ResNet-50 trained on ImageNet-1k compared to a ResNet-50 trained to mitigate texture bias \citep{geirhos_imagenet-trained_2022}. 
Since these are trained on ImageNet-1k (1 million images) rather than ImageNet-21k (14 million images), the overall performance is lower than other models we studied earlier (see Appendix \ref{app:architecture_and_mitigations}). 
Controlling for pretraining data, we observe a boost of +0.8\% in overall top-5 accuracy for the model mitigating texture bias. On the relevant subset of images (those marked with texture as a distinctive factor), we find consistent improvements in accuracy for the \texttt{low} income bucket +4.1\% and in lower performing regions (\texttt{Africa} +3.0\% and \texttt{Asia} +0.6\%).
This suggests that factor labels do not only explain model mistakes, but can also reveal potential mitigations to combat performance disparities.

\section{Related work}
\citet{rojas2022the,Singh_2022_CVPR, goyal_vision_2022, goyal_fairness_2022} have used Dollar Street to understand disparities in model performance between geographic and income groups, finding that many models perform better on images from \texttt{Europe} and \texttt{the Americas}, as well as those from higher household incomes. While many of the aforementioned works focus on how model architectures affect disparity findings, additional studies \citep{de2019does, shankar2017no} investigate the dataset itself to identify causes of variation in model performance, including the broader geographical distribution of images as compared to the model's training data. A number of datasets \citep{ramaswamy2023beyond,dubey2021adaptive,kim2021towards} and dataset auditing tools \citep{wang2022revise} have since been developed with geographic diversity in mind.
\citet{de2019does} investigated the use of English as a ``base language'' for data collection.
Empirical studies have shown that the ``concreteness'' of English words can vary greatly where crowd-sourced annotators consider words like ``human'' and ``bobsled'' more concrete than words like ``recreation'' and ``outage'' \citep{brysbaert2014concreteness}, and previous discussions of ``class label perceptions'' distinguish physical properties of a substance (such as orientation or texture) from a purpose relative to the specific being that interacts with the substance, such as being ``sit-on-able'' \citep{Gibson1979-GIBTEA}. The relationship between the true meaning of a concept versus its perceptible form remains contested for both models \citep{bender-koller-2020-climbing} and humans \citep{border}.
Beyond Dollar Street, other works study variations in representations of concepts across visual factors including pose, background, occlusion, etc. within ImageNet \citep{idrissi2022imagenet} and collect supplementary multi-class labels \citep{DBLP:journals/corr/abs-2101-05022, DBLP:journals/corr/abs-2006-07159, pmlr-v119-shankar20c}. \citet{NEURIPS2019_97af07a1} created a benchmark to measure a model's robustness to backgrounds, rotations, and viewpoints.

\section{Conclusion}
In this work, we take a step towards explaining why disparities in object-recognition systems arise. We annotate images from the Dollar Street dataset with distinguishing factors in order to explain how objects differ across incomes and geographies. Using these labels, we identify vulnerabilities in CLIP, a foundation model with impressive zero-shot classification performance. We find disparities in model performance are associated with texture, occlusion, and darker lighting.
Finally, we surface initial promising mitigations such as texture debiasing that can improve performance disparities. 
This shines light on a promising research direction leveraging techniques in robustness for fairness gains.
In future work, we plan to explore further targeted mitigations can improve performance disparities in vision systems. While our conclusions are limited by the number of samples and representative diversity of the Dollar Street dataset, we hope by releasing our factor annotations we spur further research into equitable vision systems. We have released the DollarStreet factor annotations at \url{https://github.com/facebookresearch/dollarstreet_factors}.

\paragraph{Acknowledgements}
We would like to acknowledge \textit{Adina Williams, Sahir Gomez, Somya Jain, Quentin Duval, Ida Cheng, Austin Miller} for their helpful discussions, support, and insightful feedback.

\bibliography{biblio}
\bibliographystyle{plainnat}

\newpage
\appendix
\onecolumn
\section{Appendix}
\subsection{Annotating Dollar Street with factor labels}
\label{app:annotation_all}
\subsubsection{DollarStreet Statistics}
\label{app:ds_stats}
\begin{table}[H]
    \centering
\begin{tabular}{ll|lll}
\toprule
    Region &  & & Income Level &  \\
    &  & low&  medium & high\\
\midrule
    &  Africa  & 2141 & 1443 & 280 \\
    &  Asia  & 1362 & 8673 & 1424 \\
    &  Europe & 0 & 1443 & 1455 \\
    & The Americas  & 339 & 2093 & 1223 \\
\midrule
\end{tabular}
    \caption{Number of images for each region, income level pair in Dollar Street.}
    \label{app:image_region_combination_number}
\end{table}

Table \ref{app:image_region_combination_number} shows the number of images in Dollar Street for each income and region pairing. We observe the distribution across images and regions is far from uniform, implying region and income distributions skew of counts are entangled. Consequently, we present both region and income comparisons where appropriate in our analysis.

\subsubsection{Prototypical Image Selection}
\label{app:proto}
We define prototypical images for each class as those correctly classified by ResNet-50 model with the highest confidence. We use a ResNet-50 model pre-trained on ImageNet21k from \citet{ridnik2021imagenet}. We select the ImageNet classes that overlap with Dollar Street labels, using the mapping as defined in \citep{goyal_fairness_2022}. We use a soft-max over the sub-section of ImageNet classes that are in the mapping.
We take the top predictions and confidence for these ImageNet classes and use the defined mapping from IN21k to Dollar Street in order to make DollarStreet class predictions. Out of the box, the model does not perform well on DollarStreet.
Running a full pass over the dataset with Batch Norm in train mode, without any updates to the model weights, helps with the distribution shift from ImageNet to DollarStreet images, meaning overall accuracy is higher.

We select the three images that the model predicts successfully with the highest confidence. If such images do not exist, prototypical images are hand-selected.  Table \ref{app:prototypical} shows the prototypical images used for five classes. 

\subsubsection{Annotation Setup}

\label{app:annotations}

\begin{figure}[H]
    \centering
    \includegraphics[width=\textwidth]{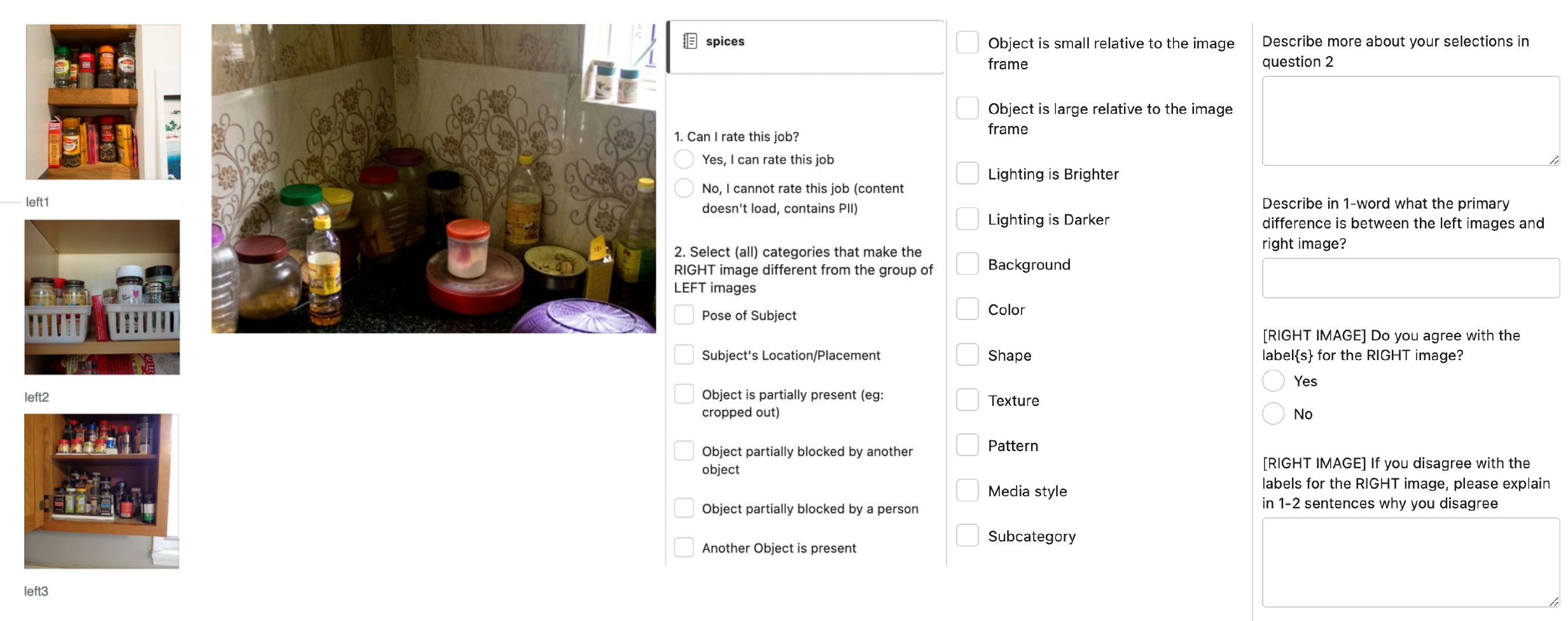}
    \vspace{-2em}
    \caption{Example annotation task.}
    \label{app:annotation_task_fig}
\end{figure}

Figure \ref{app:annotation_task_fig} shows an example of the annotation task. Annotators select the factors distinguishing each image among sixteen factors such as pose, various forms of occlusion, size, style, type or breed. 
Annotators can select any number of distinctive factors for each image. 
We source 10 annotators through a third party vendor from South East Asia.
In addition, we ask annotators to provide text descriptions to account for factors outside the sixteen we provide. We trained annotators with examples so that they were familiar with the task before annotating the target images. We had intermediate QA from the third party vendor monitoring annotations for quality. We also ask annotators whether they agree with the original class label for each image.
\vspace{0em}

\begin{table}[H]
\centering
\begin{tabular}{r | c   c  c } 
Class & & Prototypical Images &  \\
\hline
\vspace{-1em}
\\
grains & \includegraphics[width = 0.1\textwidth]{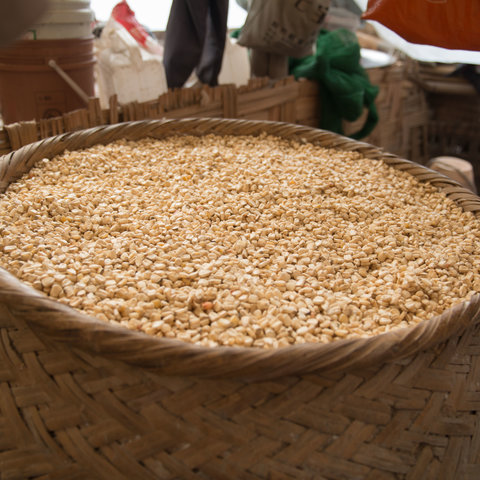} & \includegraphics[width = 0.1\textwidth]{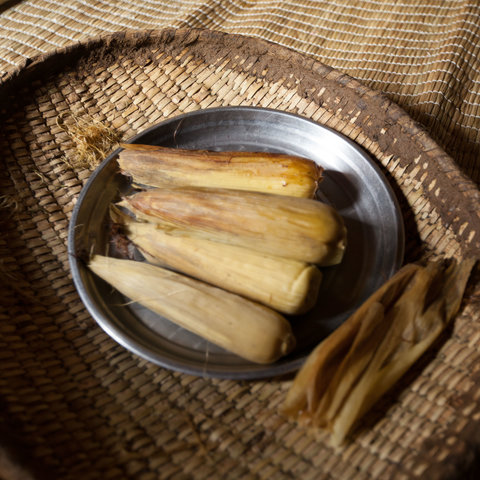} & \includegraphics[width = 0.1\textwidth]{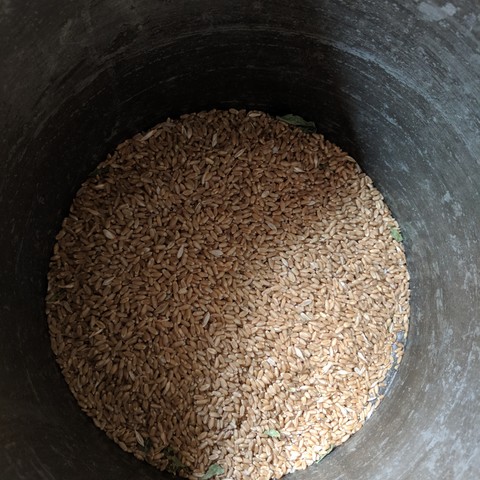} \\ 
 \hline 
 \\
plates & \includegraphics[width = 0.1\textwidth]{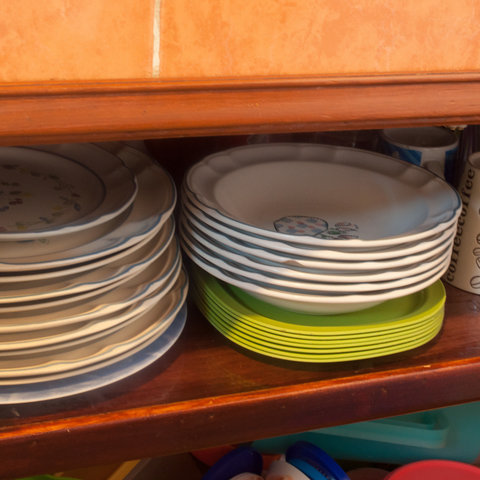} & \includegraphics[width = 0.1\textwidth]{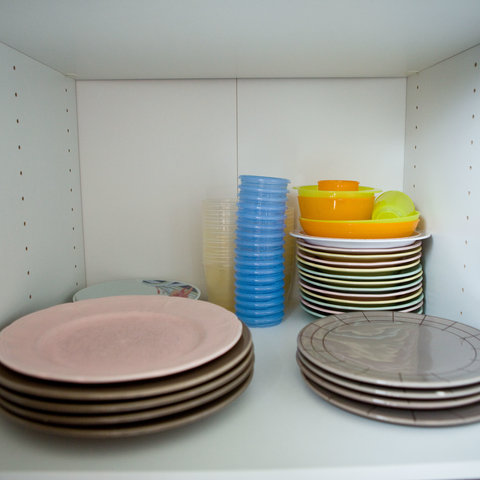} & \includegraphics[width = 0.1\textwidth]{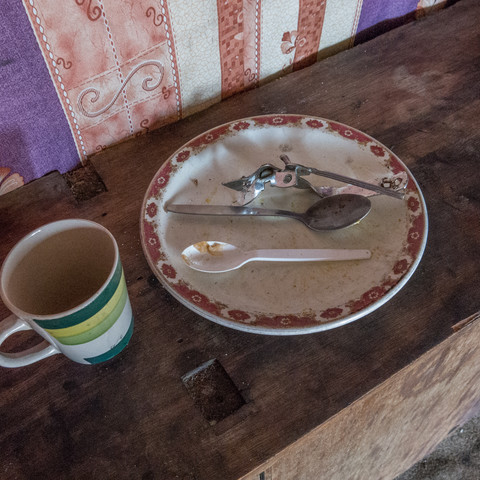} \\ 
 \hline  \\
power outlets & \includegraphics[width = 0.1\textwidth]{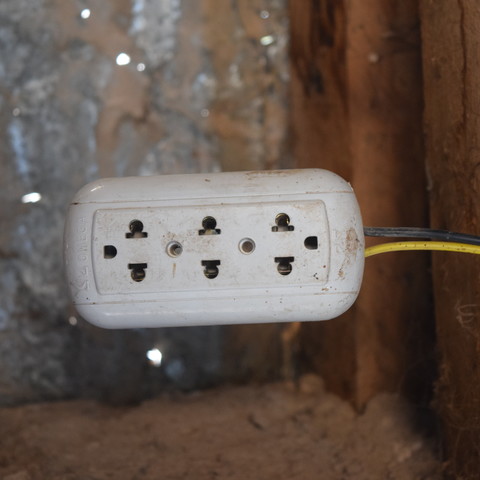} & \includegraphics[width = 0.1\textwidth]{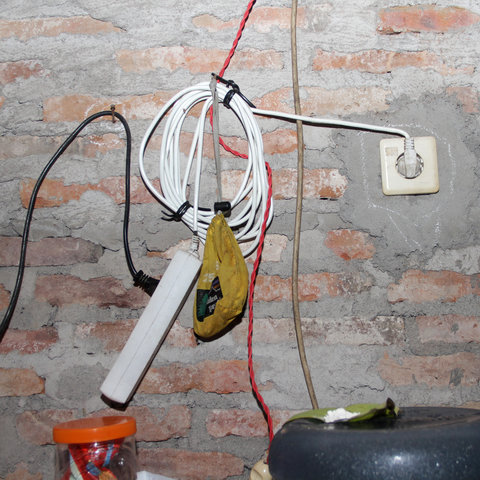} & \includegraphics[width = 0.1\textwidth]{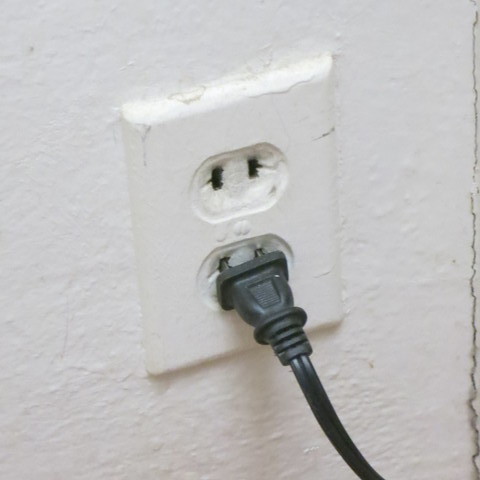} \\ 
 \hline \\
cleaning floors & \includegraphics[width = 0.1\textwidth]{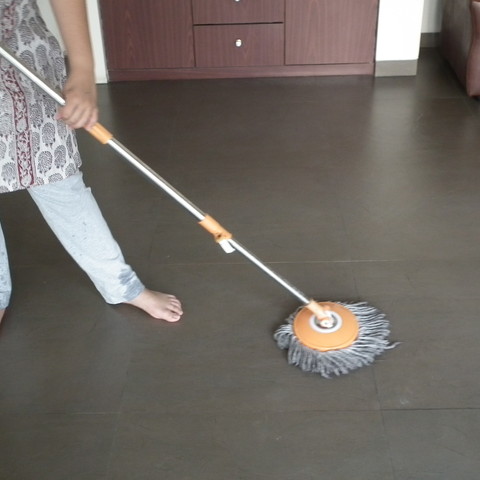} & \includegraphics[width = 0.1\textwidth]{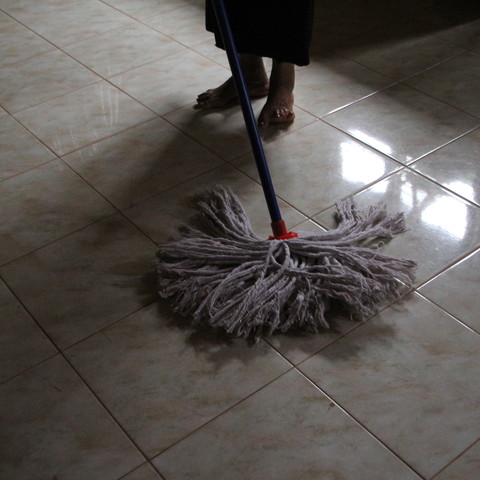} & \includegraphics[width = 0.1\textwidth]{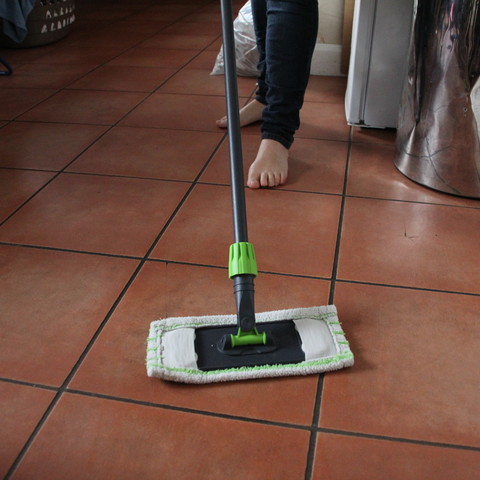} \\ 
 \hline \\
toothbrushes & \includegraphics[width = 0.1\textwidth]{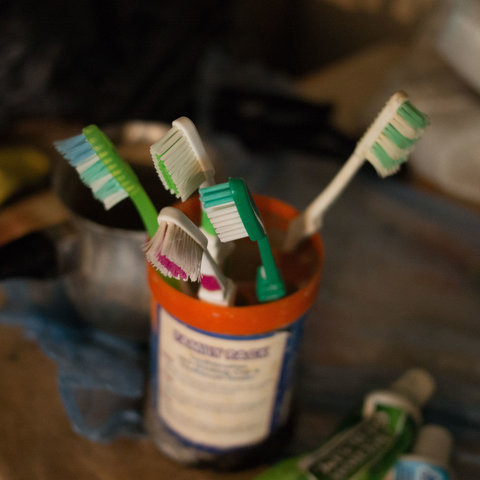} & \includegraphics[width = 0.1\textwidth]{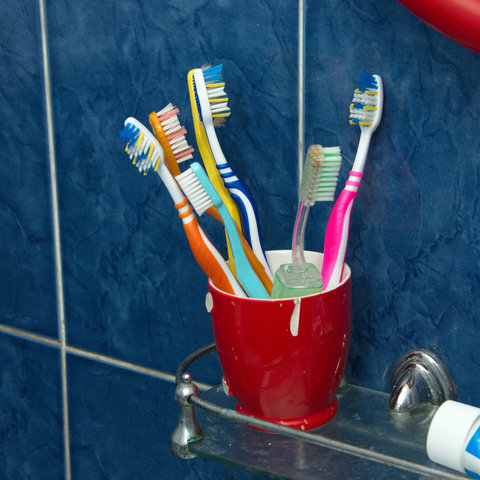} & \includegraphics[width = 0.1\textwidth]{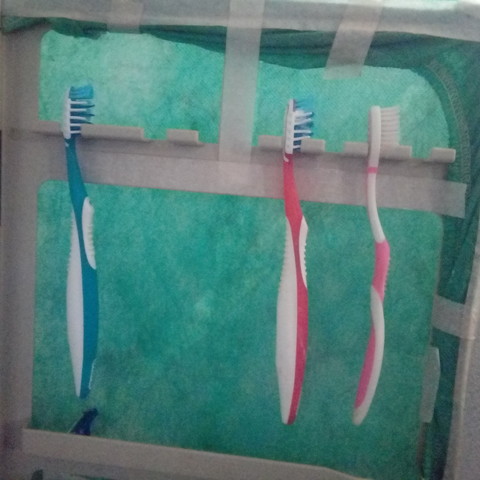} \\ 
\hline
 
\end{tabular} 
\caption{Prototypical images used for five classes.} 
\label{app:prototypical}
\end{table}

\subsubsection{Label Agreement Annotation Setup}
\label{app:label_agreement_overveiw}
\begin{figure}
    \centering
    \includegraphics[width=\textwidth]{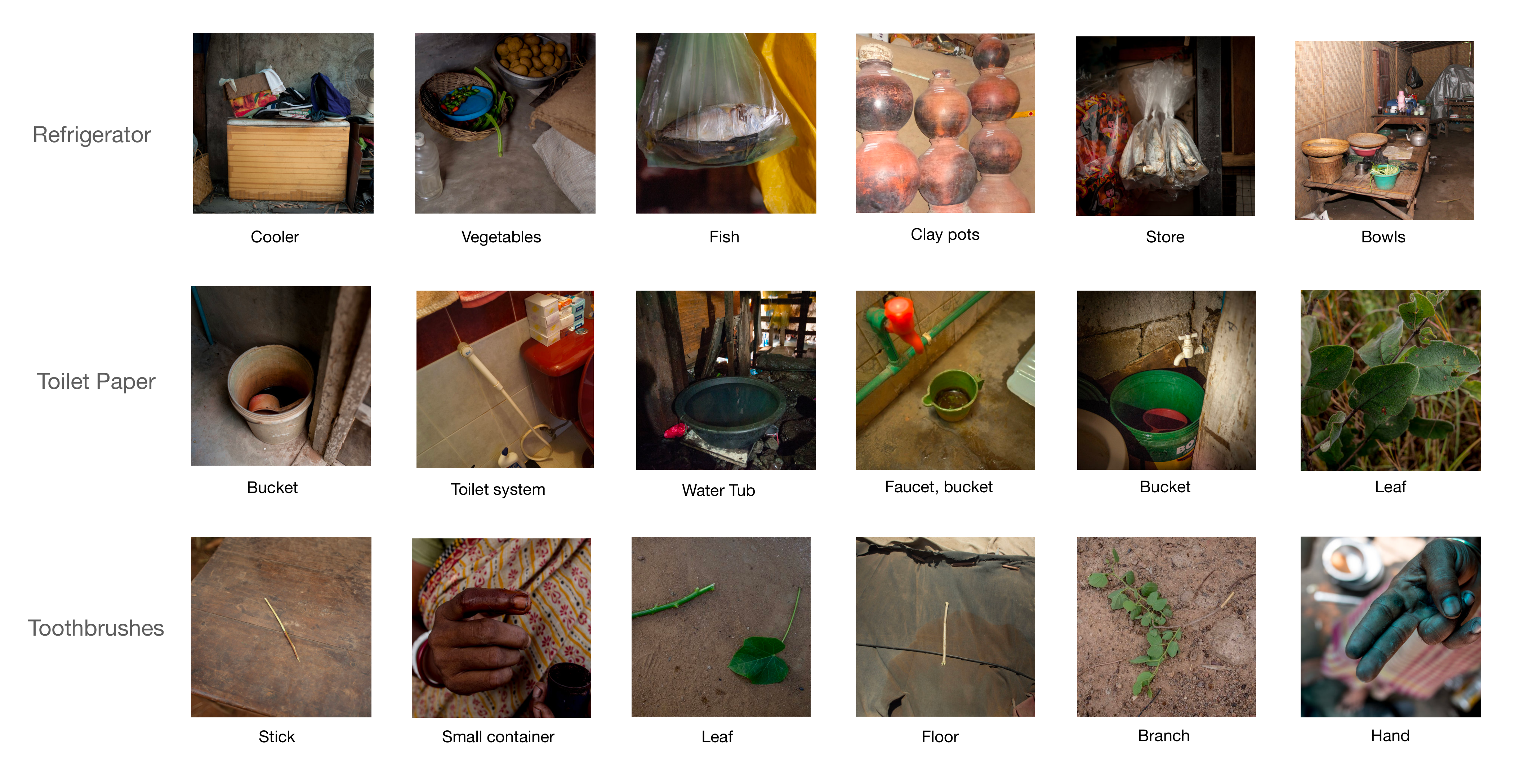}
    \caption{Randomly sampled example images and alternative labels given for the three classes with most disagreement. The original class label is shown on the left, and the alternative label given by the annotator shown below each image.}
    \label{fig:image_examples_affordances}
\end{figure}

\begin{table}[H]
    \centering
    \begin{tabular}{l l}
    \toprule
               Country & Number of annotators \\
    \midrule
     India & 8\\
     Nigeria & 9 \\
     Brazil & 13 \\
     United Arab Emirates & 6 \\
     United States & 8\\
    \bottomrule
    \end{tabular}
    \label{table_annotators2}
    \caption{Annotator breakdown for label agreement task.}
\end{table}
For our follow up annotations about label agreement, we sourced 44 annotators from 5 different countries, with the full demographics shown in Table \ref{table_annotators2}. We asked one annotator per country about each image in question. In Table \ref{fig:image_examples_affordances}, we show example images from the three most disputed classes, along with alternative labels suggested by annotators.  In Table \ref{tab:top_disagree_class}, we show the classes with the highest and lowest levels of disagreement among annotators. 

\begin{table}
\centering
\begin{tabular}{lll}
\toprule
           class &                 \% disagreement      \\
\midrule
 toilet paper &              88.4 \\
 refrigerators &       83.5 \\
  toothbrushes & 79.3 \\
sofas &   77.8 \\
diapers &         72.0 \\
 armchairs &       70.6  \\
    showers &       66.3  \\
 kitchen sinks &         64.5  \\
  wall clocks &       63.2  \\
    radios &         60.4 \\
\bottomrule
\end{tabular}
\quad
\begin{tabular}{lll}
\toprule
           class &                 \% disagreement      \\
\midrule
 medication &              10.0 \\
 fruit trees &       11.8 \\
  plates of food & 12.3 \\
trash &   14.3 \\
cleaning floors &         14.5 \\
 ceilings &       15.0  \\
    homes &       15.0  \\
 books &         15.0  \\
  cooking pots &       19.8  \\
    wheel barrows &       20.0 \\
\bottomrule
\end{tabular}

    \caption{Top ten classes with the highest percentage of annotators who \emph{disagreed} (left) and \emph{agreed} (right) with the original class label}
    \label{tab:top_disagree_class}
\end{table}

\subsection{How do objects vary across incomes and geographies?}
\label{app:object_variation}
We show the most dissimilar classes across incomes and regions by comparing the Jensen-Shannon Distance of the factor annotation distributions in Tables \ref{app:dissimilar_class_across_incomes} and \ref{app:dissimilar_class_across_regions}.

\begin{table}[H]
    \centering
\begin{tabular}{lll}
\toprule
               class & income bucket &              differentiating factors \\
\midrule
               roofs &   \texttt{low} vs. \texttt{high}&       subcategory, pose, smaller \\
            ceilings &   \texttt{low} vs. \texttt{high}&       pose, subcategory, texture \\
             diapers &   \texttt{low} vs. \texttt{high}&            color, shape, texture \\
              radios &   \texttt{low} vs. \texttt{high}&        color, shape, subcategory \\
              floors &   \texttt{low} vs. \texttt{high}&           texture, pose, pattern \\
               sofas &   \texttt{low} vs. \texttt{high}& color, texture, multiple objects \\
       kitchen sinks &   \texttt{low} vs. \texttt{high}&          shape, pose, background \\
        toilet paper &   \texttt{low} vs. \texttt{high}&          color, pose, background \\
           wardrobes &   \texttt{low} vs. \texttt{high}&       background, pattern, color \\
mosquito protections &   \texttt{low} vs. \texttt{high}&      color, subcategory, pattern \\
\bottomrule
\end{tabular}
    \caption{Classes with most stark differences in factor distributions by Jensen-Shannon Distance (JSD) across incomes.}
    \label{app:dissimilar_class_across_incomes}
\end{table}

\begin{table}[H]
    \centering
\begin{tabular}{lll}
\toprule
           class &                 regions &              distinctive factors \\
\midrule
 chickens &          \texttt{Asia} vs.  \texttt{Europe} &         partial view, color, shape \\
 chickens &        \texttt{Europe} vs.  \texttt{Africa} &          pose, partial view, color \\
  diapers & \texttt{The Americas} vs.  \texttt{Africa} &          pose, color, partial view \\
pet foods &    \texttt{Asia} vs. \texttt{The Americas} &            color, texture, pattern \\
pet foods &          \texttt{Asia} vs.  \texttt{Europe} &        pattern, subcategory, color \\
 ceilings &        \texttt{Europe} vs.  \texttt{Africa} &         pose, subcategory, texture \\
    roofs &        \texttt{Europe} vs.  \texttt{Africa} &         subcategory, pose, texture \\
 car keys &          \texttt{Asia} vs.  \texttt{Europe} & pattern, partial view, subcategory \\
  make up &        \texttt{Europe} vs.  \texttt{Africa} &   background, subcategory, pattern \\
    goats &          \texttt{Asia} vs.  \texttt{Africa} &        pattern, color, subcategory \\
\bottomrule
\end{tabular}

    \caption{Classes with most stark differences in factor distributions by Jensen-Shannon Distance (JSD) across regions}
    \label{app:dissimilar_class_across_regions}
\end{table}

We show the most similar classes across incomes and regions using the same procedure of comparing Jensen-Shannon Distance of the factor annotation distributions in Tables \ref{app:similar_class_across_incomes} and \ref{app:similar_class_across_regions}.

\begin{table}[H]
    \centering
\begin{tabular}{lll}
\toprule
                  class &  income buckets &                 distinctive factors \\
\midrule
        vegetable plots &    \texttt{low} vs. \texttt{high}& multiple objects, background, color \\
                 phones &  \texttt{medium} vs. \texttt{high}&  background, pose, multiple objects \\
                   pens &  \texttt{medium} vs. \texttt{high}&          color, background, pattern \\
                  bikes &    \texttt{low} vs. \texttt{high}&    background, subcategory, smaller \\
              armchairs &  \texttt{medium} vs. \texttt{high}&             color, background, pose \\
latest furniture bought &  \texttt{medium} vs. \texttt{high}&      subcategory, background, color \\
            child rooms &  \texttt{medium} vs. \texttt{high}&                pose, pattern, color \\
            wall clocks &  \texttt{medium} vs. \texttt{high}&                  color, pose, shape \\
       cooking utensils &  \texttt{medium} vs. \texttt{high}&                pose, shape, pattern \\
\bottomrule
\end{tabular}
    \caption{Classes most similar in factor distributions by Jensen Shannon Distance across incomes}
    \label{app:similar_class_across_incomes}
\end{table}

\begin{table}[H]
    \centering
\begin{tabular}{lll}
\toprule
            class &                 regions &                   distinctive factors \\
\midrule
  vegetable plots & \texttt{The Americas} vs.  \texttt{Africa} &             pose, background, pattern \\
           phones &          \texttt{Asia} vs.  \texttt{Europe} &               pose, background, color \\
             pens &        \texttt{Europe} vs.  \texttt{Africa} &                  pose, color, pattern \\
    wheel barrows &  \texttt{Europe} vs. \texttt{The Americas} &               color, pose, background \\
         ceilings &          \texttt{Asia} vs.  \texttt{Africa} &         subcategory, pattern, texture \\
             pets &          \texttt{Asia} vs.  \texttt{Europe} &      background, pattern, subcategory \\
           stoves &          \texttt{Asia} vs.  \texttt{Africa} &           subcategory, color, pattern \\
menstruation pads &    \texttt{Asia} vs. \texttt{The Americas} &            pose, subcategory, pattern \\
              tvs &  \texttt{Europe} vs. \texttt{The Americas} & partial view, subcategory, background \\
   everyday shoes &  \texttt{Europe} vs. \texttt{The Americas} &            color, partial view, shape \\
\bottomrule
\end{tabular}
    \caption{Classes most similar in factor distributions by Jensen Shannon Distance (JSD) across regions}
    \label{app:similar_class_across_regions}
\end{table}

\subsection{Evaluation Setup}
\label{app:eval_setup}
\paragraph{CLIP Prompt Engineering} We use CLIP in a zero shot setting, where we prompt the model using the set of Dollar Street classes (e.g. \textit{medication, plates of food}) for each image to generate predictions. We generate the text prompts for CLIP by combining the 80 prompt templates used in the original CLIP paper with each Dollar Street class name, substituting \texttt{\_} for spaces. We consider an image correctly predicted if the top 5 classes predicted by CLIP is associated with the photo. \textit{Note: Most photos in DollarStreet have only one label, but a small subset of (638) images
containing multiple class labels (e.g. (cups, plates, dish racks) and (child rooms,
kids bed, beds))}. 

\paragraph{ImageNet21k as a shared taxonomy} For models outside of CLIP, we use ImageNet21k to ground our models in a shared taxonomy. Following \citet{goyal_fairness_2022}, we map the ImageNet21k labels to DollarStreet classes. We consider the image correctly classified if any of the top 5 ImageNet21k classes predicted by the model are mapped to any of the DollarStreet classes associated with the photo. We note that the mapping is not 1:1, and multiple classes in DollarStreet have multiple classes in ImageNet 21k that map to the single class. All of the models used for evaluation excluding CLIP and SEER are trained on ImageNet 21k. SEER is pre-trained in a self-supervised manner, and the model is fine-tuned on the 108 classes in ImageNet 21k that overlap with DollarStreet prior to evaluation.
For ImageNet-21k pretraining, we use models from \citet{ridnik2021imagenet}.

\label{app_sec:worst_classes}
\paragraph{Class level performance disparities}
\begin{figure}[H]
\centering
\begin{subfigure}{}
\includegraphics[width=0.7\textwidth]{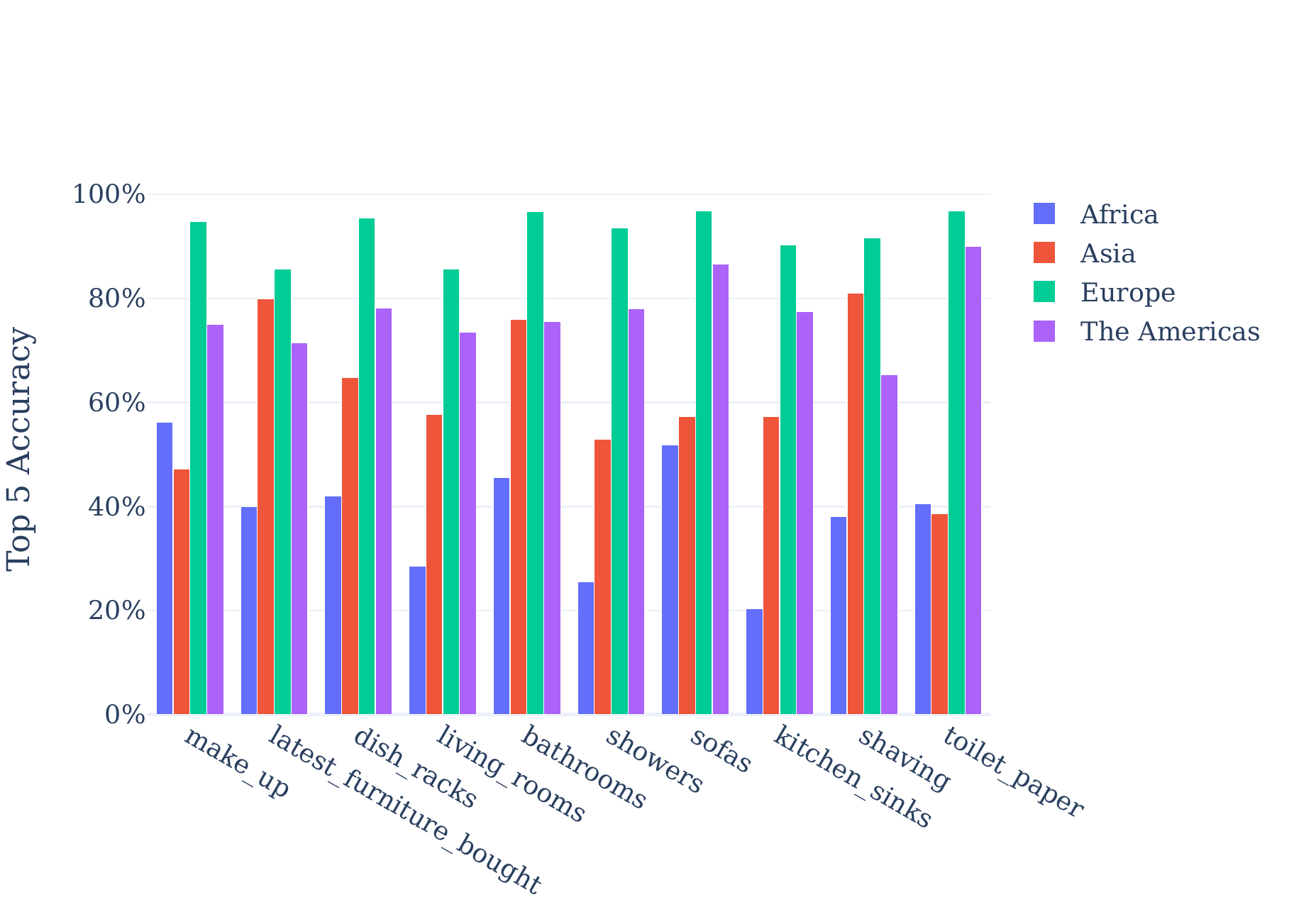} \end{subfigure}
\begin{subfigure}{}
\includegraphics[width=0.7\textwidth]{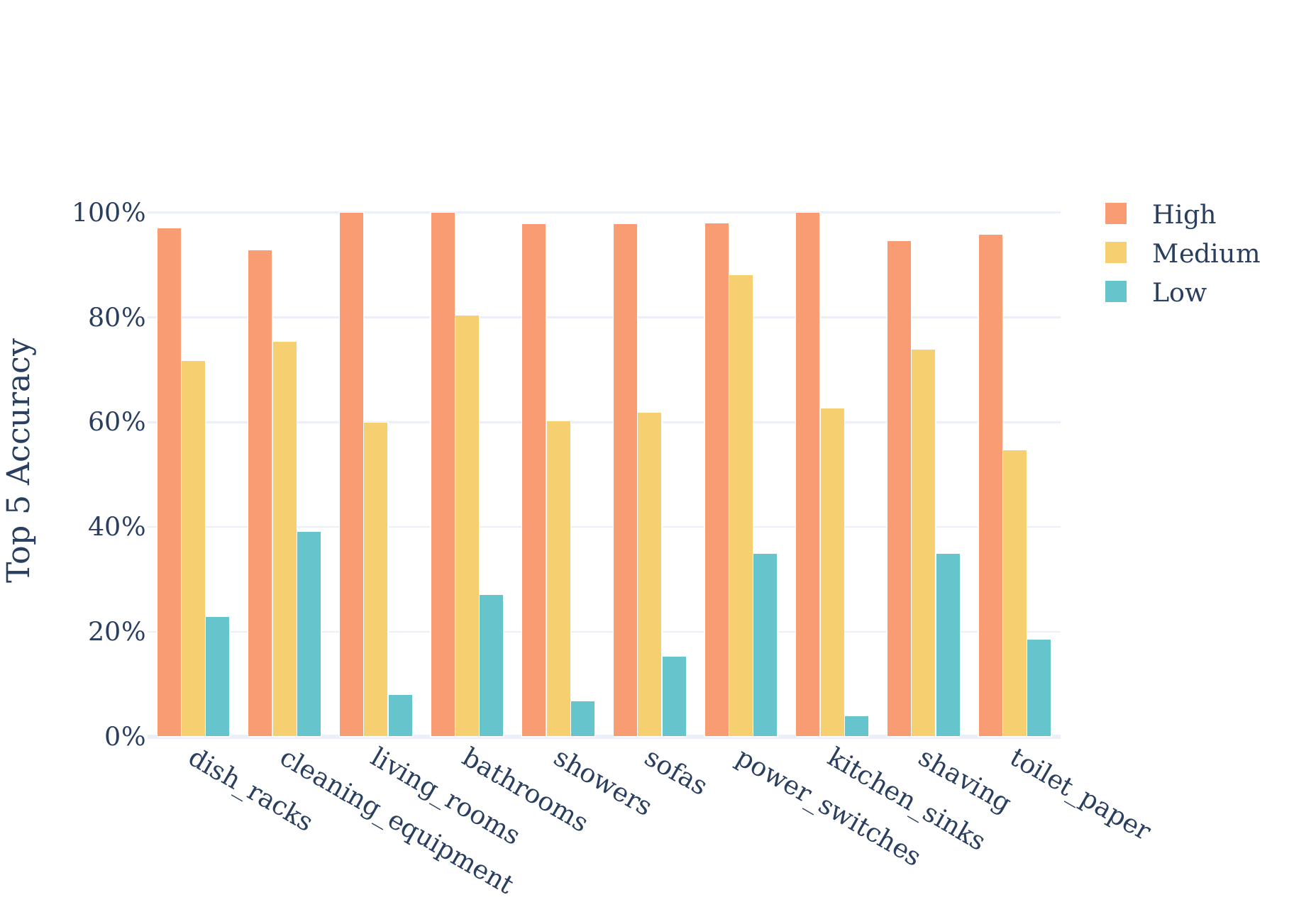} \end{subfigure}
\caption{10 classes with biggest performance discrepancy over regions (top) and income bucket (bottom).}
\label{fig:app_class_level_acc}
\end{figure}

Figure \ref{fig:app_class_level_acc} and show the top 10 classes with the biggest performance disparity between groups for regions and incomes. We define the largest performance discrepancy as the maximum difference in accuracy between any two regions (or income buckets). At a class level, we find that the discrepancy in accuracy can be stark - over 50\% for the classes with the widest gap. For both incomes and geographies, we find that the differences mostly pertain to items in kitchens (\textit{dish racks, kitchen sinks}) and items in bathrooms (\textit{showers, shaving, toilet paper, bathrooms}).

\subsection{Explaining model performance disparities with factor labels}
\label{app:explain_model_mistakes}

As part of our analysis of model performance disparities, we investigate the impact of pretraining class balance and image quality. In Table \ref{tab:pearson_corr_quality}, we show the Pearson correlation coefficients and p-values between each model's top-5 accuracy and the Image DPI, a measure of image resolution. In Table \ref{tab:pearson_corr_class_balance}, we show the Pearson correlation coefficients and p-values between each model's top-5 accuracy and the ImageNet-21K class count. We excluded CLIP from this analysis as CLIP was trained on a proprietary dataset. 

\begin{table}[H]
\centering
\begin{tabular}{lll}
\toprule
           Model & \makecell{Correlation Between \\ Top 5 Accuracy and Image DPI}  \\
\midrule
 ViT &  -0.019 (p = 0.035) \\
 ResNet50 &  -0.023 (p = 0.008)  \\
MLPMixer & -0.026 (p = 0.002)  \\
BeIT &   -0.003  (p = 0.72) \\
SEER &    -0.016 (p = 0.057)  \\
 CLIP &    -0.035 (p = 0.00005)  \\
\bottomrule
\end{tabular}

\caption{Pearson Correlation coefficients and p-values between each model's top-5 accuracy and image quality, as measured by DPI.}
\label{tab:pearson_corr_quality}
\end{table}

\begin{table}[H]
\centering
\begin{tabular}{lll}
\toprule
Model &  \makecell{Correlation Between \\Top-5 Accuracy and Class Count}  \\
\midrule
ViT &  0.126 (p $<$ 0.0001) \\
ResNet50 &  0.142 (p $<$ 0.0001)  \\
MLPMixer & 0.135 (p $<$ 0.0001)  \\
BeIT &  0.222 (p $<$ 0.0001) \\
SEER &  0.103 (p $<$ 0.0001)  \\
\bottomrule
\end{tabular}

\caption{Pearson Correlation coefficients and p-values between each model's top-5 accuracy and ImageNet-21K class counts. CLIP is not included, as it was trained on a proprietary dataset.}
\label{tab:pearson_corr_class_balance}
\end{table}

Factors most associated with misclassifications differ considerably across regions and incomes. 
We find  for the \texttt{high} income bucket, 
objects marked as \textit{smaller} are most associated with mistakes, appearing +2.8x more among mistakes.
On the other hand, \textit{texture} which is not among the top five factors among mistakes in the \texttt{high} income bucket is associated
with mistakes in the \texttt{medium} and \texttt{low} income buckets. 
\textit{Texture} is +0.6x and +1.7x more likely to appear among mistakes in the  \texttt{medium} and \texttt{low} income buckets respectively.
We also find in the \texttt{low} income bucket, factors such as \textit{occlusion} and \textit{darker lighting} to be associated with model mistakes, appearing 
+1.2x and +0.9x more so among mistakes in the \texttt{low} income bucket. 
This suggests specific factors such as \textit{texture}, \textit{occlusion}, and \textit{darker lighting} are associated with the disparity in performance we observe across incomes.
\vspace{-2em}
\begin{figure}[H]
\centering
\begin{subfigure}{}
\includegraphics[width=0.7\textwidth]{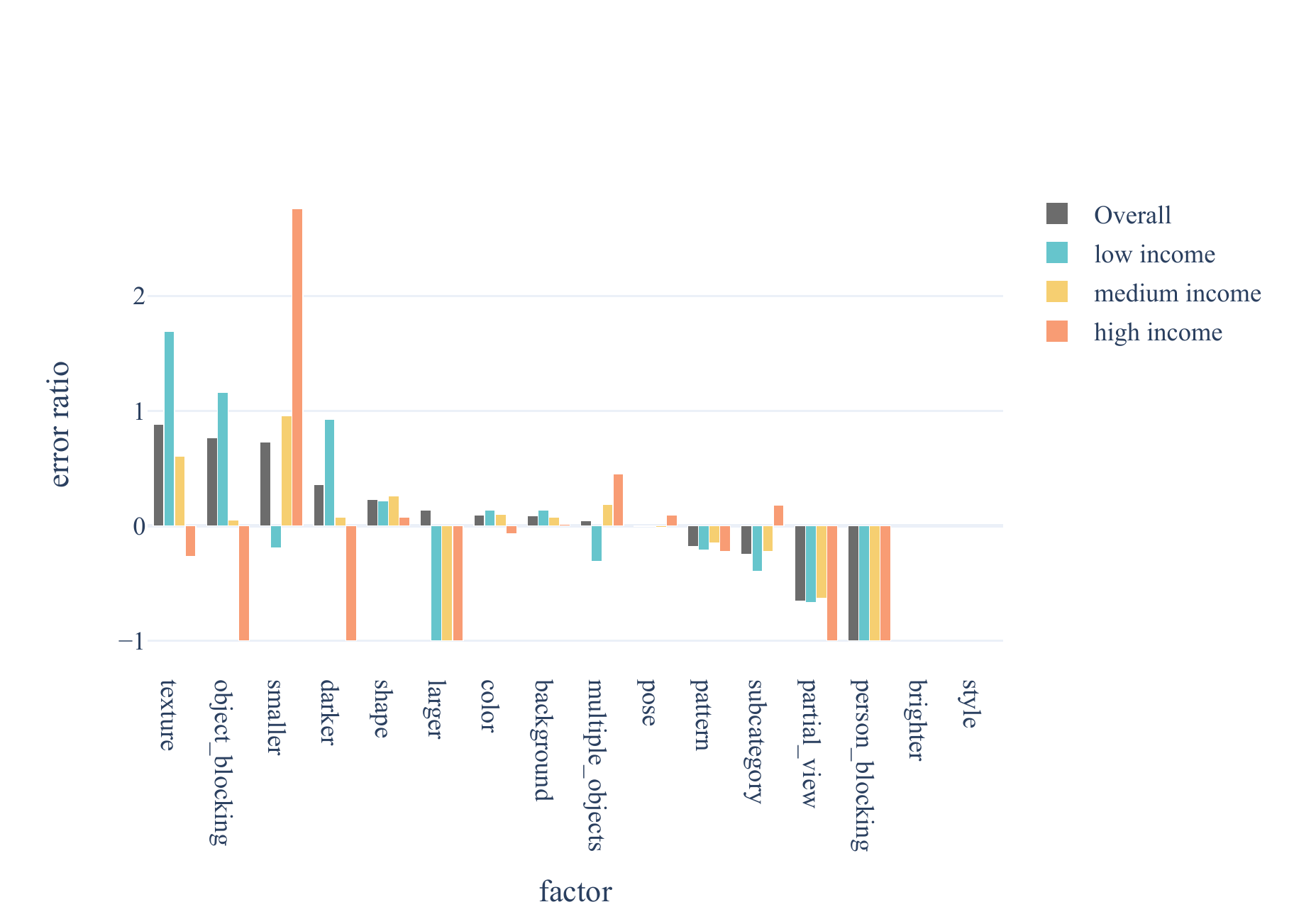}
             \end{subfigure}
\vspace{-2em}
\begin{subfigure}{}
\includegraphics[width=0.7\textwidth]{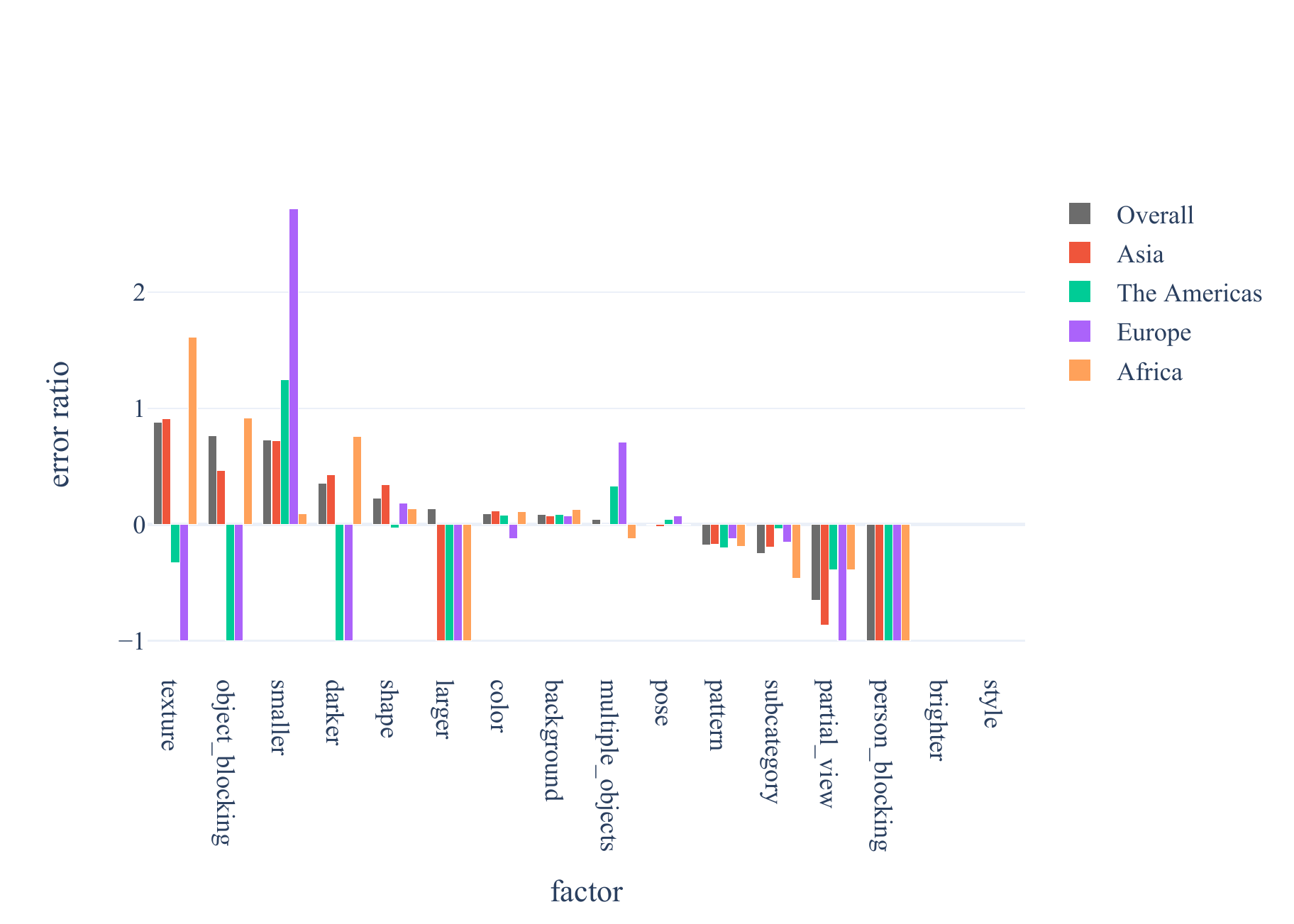}
        \end{subfigure}
\caption{Shows the full error ratios for each factor per income bucket (top) and region (bottom). An error ratio higher than zero indicates the factor is more associated with model mistakes; less than zero indicates the factor is less likely to appear among a model's mistakes.}
\label{app_fig:factor_per_region_inc}
\end{figure}

\paragraph{Factors associated with mistakes differ across regions.}
We also measure the factors associated with model mistakes across regions in Figure \ref{fig:top_5_factors_per_region}.
In  \texttt{Asia} we observe the factors most associated with mistakes are similar to those associated with mistakes overall.
However, we find distinctive factors are associated with mistakes across each of the other regions.
In the Americas, we find \textit{smaller objects} (+1.2x  more likely to appear among mistakes), followed by images with \textit{multiple objects} (+0.3x).
Similarly in \texttt{ \texttt{Europe}}, \textit{smaller objects} and \textit{multiple objects} are most associated with mistakes appearing +2.8x and +0.7x more so among mistakes respectively.
In  \texttt{Africa} however, we find instead \textit{texture} (+1.6x) most associated with mistakes, followed by \textit{occlusion} (+0.9x) and \textit{darker lighting} (+0.8x).
This suggests the disparity due to lower performance in regions such as  \texttt{Africa} are associated with distinct factors related to \textit{texture}, \textit{occlusion}, and \textit{darker lighting}.

\paragraph{Statistical significance of error ratios for top factors.} To confirm the top factors associated with model mistakes measured by our error ratio are statistically significant. We conduct a Chi-Squared test comparing the overall distribution of counts of the top factors to their distribution of counts among misclassifications. We find a statistically significant difference with a Chi-Squared statistic of 21.7 (p-value =0.0002).

\paragraph{Factors most associated with largest discrepancies for classes across income buckets.}
We show the three factors most associated with model mistakes for the classes across income buckets with largest performance gap in Table \ref{tab:worst_class_income_bucket_explanation}. 
Trends are similar to those shown in the main paper for the largest disparity per region.

\begin{table}[H]
\centering
\label{tab:worst_class_income_bucket_explanation}
\begin{tabular}{lll}
\toprule
        Class & Income &                   Factors associated with mistakes \\
\midrule
        sofas &    \texttt{low}&  pattern (+0.5x), background (+0.3x), pose (+0.2x) \\
 toilet paper &    \texttt{low}&      texture (+3.3x), shape (+2.7x), color (+0.8x) \\
 living rooms &    \texttt{low}&  background (+0.8x), pose (+0.0x), color (-0.1x) \\
kitchen sinks &    \texttt{low}&    color (+0.5x), background (+0.3x), pose (+0.2x) \\
      showers &    \texttt{low}& background (+0.9x), pose (+0.3x), pattern (-0.5x) \\
\bottomrule
\end{tabular}
\caption{
Class-specific vulnerabilities surfaced by our factor labels.
We show vulnerabilities for the classes with lowest income performance.The values in parenthesis 
                    indicate how much more likely a factor is to appear for misclassified samples.}
\end{table}

\subsection{The effect of architecture and training procedure}
\label{app:architecture_and_mitigations}

\paragraph{Texture debiasing experimental details}
To measure the effect of reducing texture bias from \citet{geirhos_imagenet-trained_2022}, we create a mapping from Dollar Street classes to ImageNet-1k similar to \citet{rojas2022the}.
We initialize the mapping by matching the embedding similarity of each class name to its nearest neighbors from ImageNet-1k using a pre-trained Spacy language model \texttt{eng-large} \url{https://spacy.io/usage/linguistic-features#vectors-similarity}. We then manually correct any issues in this mapping to produce ImageNet-1k mappings for approximately half of the Dollar Street classes. Note for all other analysis we use the ImageNet-21k mapping from \citet{goyal_fairness_2022}. 

\paragraph{Additional experiment: choice of encoder architecture can also improve performance discrepancy}
We found CLIP with a ViT encoder has a smaller error ratio for objects marked as \emph{larger} compared to CLIP with a ResNet-101 encoder in Figure \ref{fig:top5_factors_per_model}. 
Does this insight imply improved performance robustness to objects associated with the larger factor? 
If so, we expect images annotated with the larger factor to show improved performance for the ViT encoder.
We compare the top-5 accuracy of the ViT versus ResNet encoder for images marked with larger factor. 
This suggests modeling choices such as architecture are potential mitigations to improve robustness to certain factors.
We find ViT classification accuracy is 80.4\% compared to ResNet encoder's 73.9\% suggesting the improved factor error ratio of the ViT encoder to larger objects yields a gain in performance.
For images marked as \emph{larger}, can  such a performance improvement also improve the disparity we observe across incomes?
Although the number of samples is limited ($\sim50$ images), we find ViT's improved robustness to larger objects decreases the gap between \texttt{low} and \texttt{high} income buckets for those images by 15\% compared to the ResNet encoder.
This not only suggests choices such as architecture can influence the model disparities we observe, but that the factor annotations can help uncover 
mitigations to certain factors associated mistakes to improve the disparities in performance across incomes and geographies.
This preliminary finding offers an encouraging direction to improve the model disparities we observe through the labeled factors.

\subsection{Interactive factor dashboard}

\begin{figure}[H]
\caption{Interactive dashboard for Dollar Street factor annotations with an income and factor label query (for texture).}
\centering
\includegraphics[width=\textwidth]{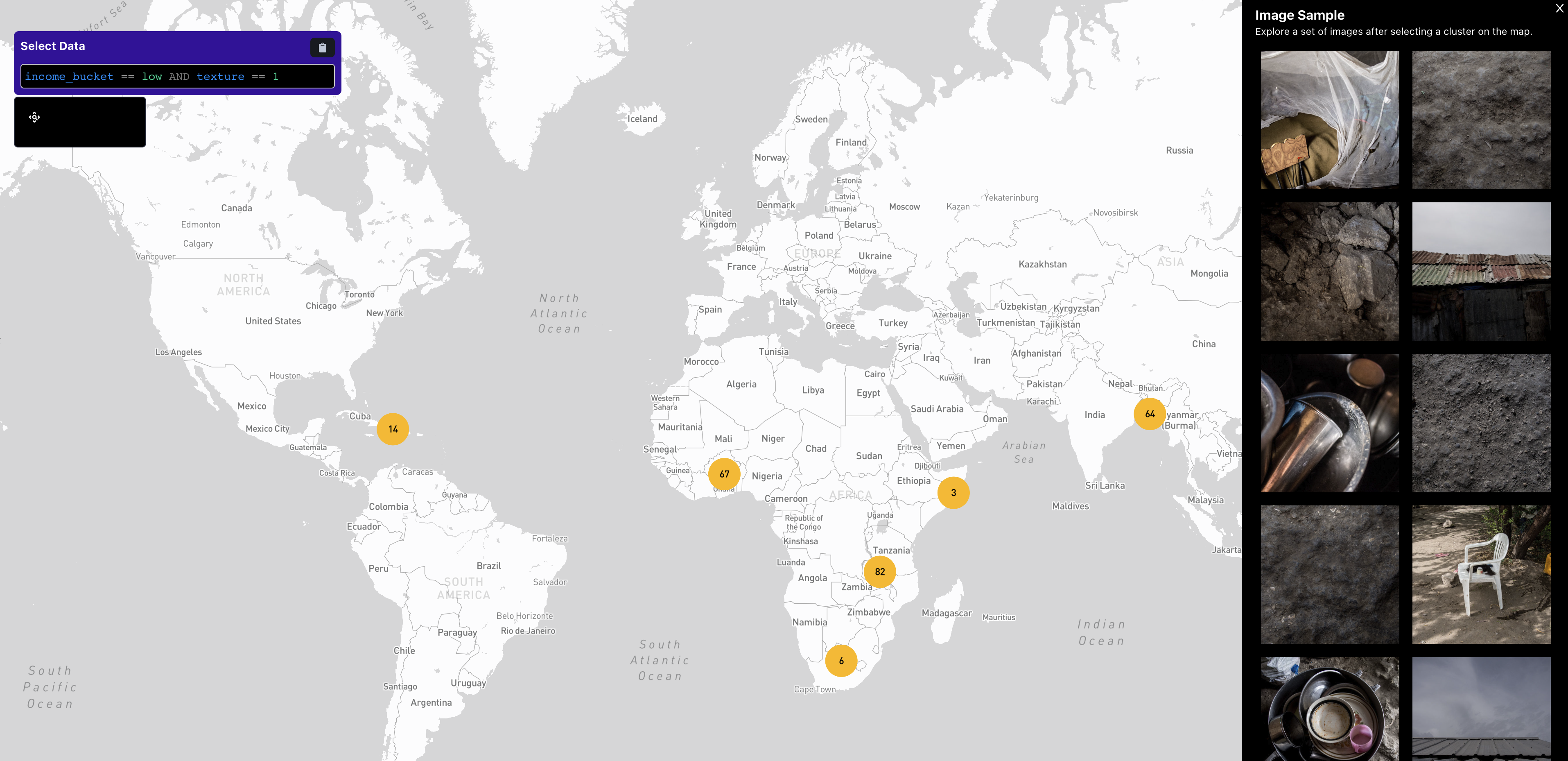}
\label{app_fig:dash1}
\end{figure}

\begin{figure}[H]
\caption{Interactive dashboard for Dollar Street factor annotations illustrating an example of the annotations.}
\centering
\includegraphics[width=\textwidth]{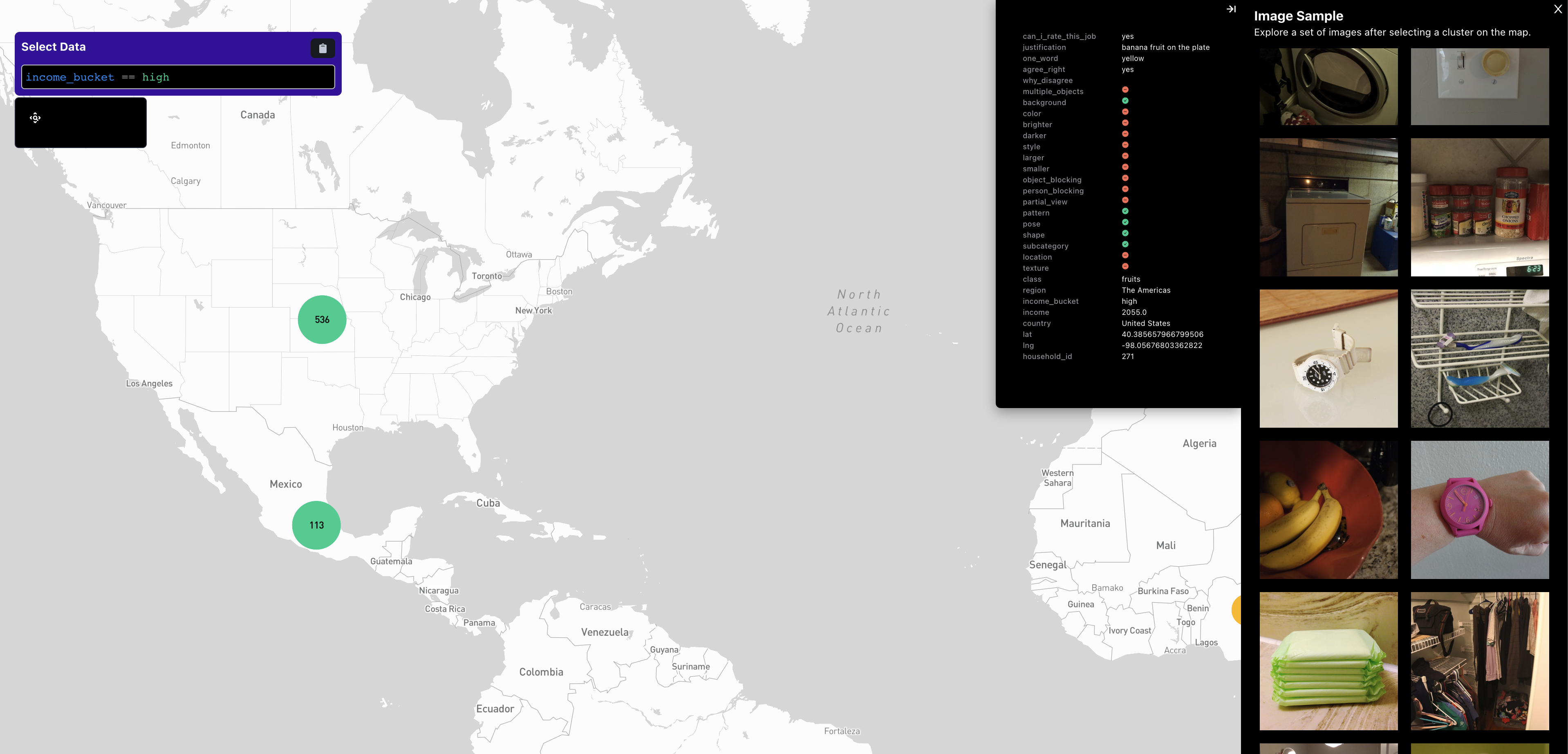}
\label{app_fig:dash2}
\end{figure}

We show screenshots of our interactive dashboard for exploring the factor labels across regions in Figures \ref{app_fig:dash1} and \ref{app_fig:dash2}. The dashboard allows for interactive queries by region, income, factor label. 
Each query yields sample images, which you can interactively explore annotations for as shown in \ref{app_fig:dash2}. We hope this tool will allow researchers to easily explore factor labels associated with images across axes such as regions or incomes to spur further research into reliable vision systems.

\subsection{Sample images}
\label{app:samples}
\begin{center}
\begin{table}[H] 
\begin{tabular}{c | c  c c  c } 
Country & & & &  \\
\hline
The Americas & \includegraphics[width = 0.15\textwidth]{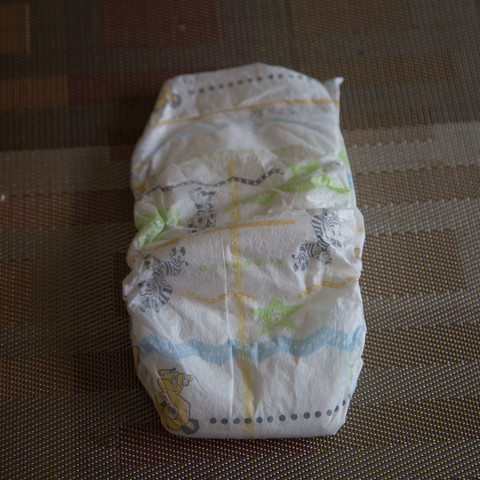} & \includegraphics[width = 0.15\textwidth]{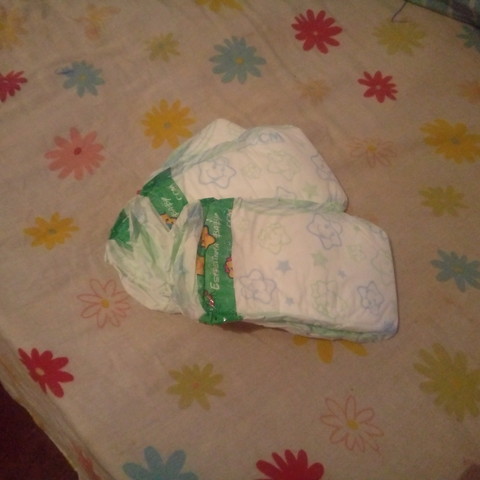} & \includegraphics[width = 0.15\textwidth]{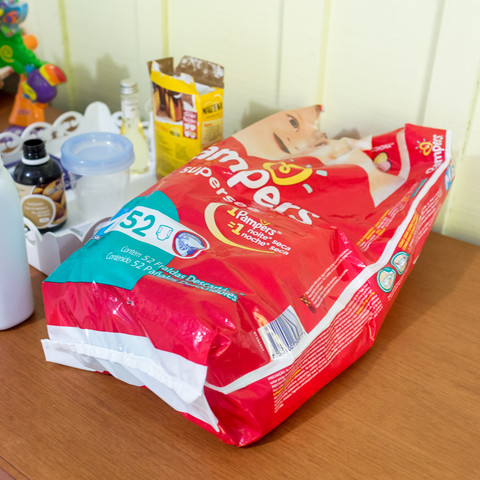} & \includegraphics[width = 0.15\textwidth]{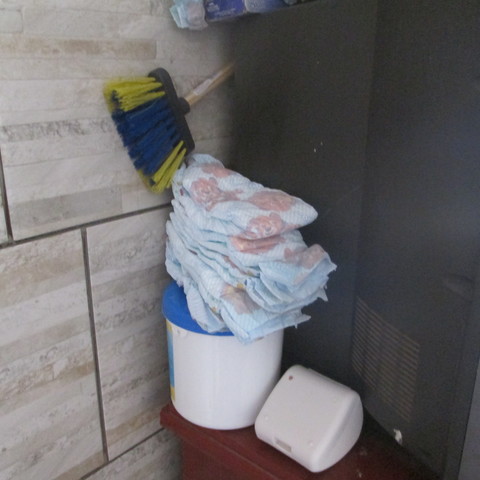} \\ 
 \hline 
 \\Africa & \includegraphics[width = 0.15\textwidth]{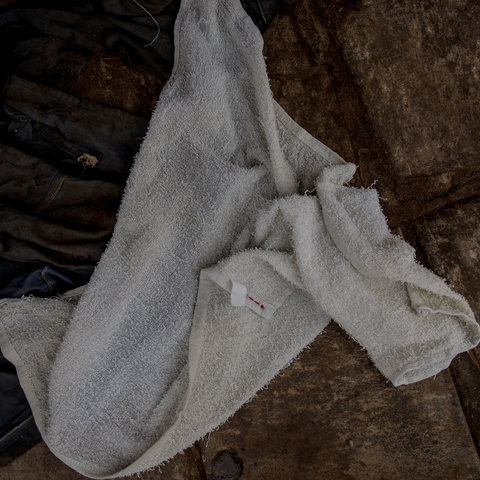} & \includegraphics[width = 0.15\textwidth]{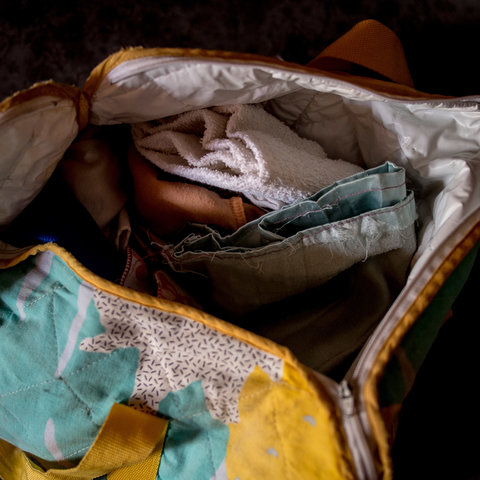} & \includegraphics[width = 0.15\textwidth]{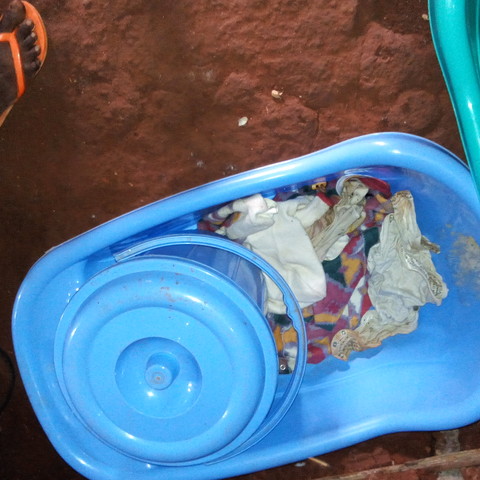} & \includegraphics[width = 0.15\textwidth]{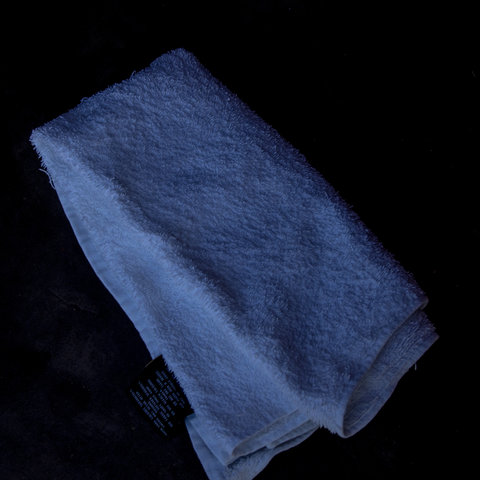} \\ 
 \hline 
\end{tabular} 
\label{app:diapers_jsd}
\caption{Examples of diaper images. Our factors surfaced that images of diapers in Dollar Street between regions differed most among \textit{pose}, \textit{color}, \textit{partial view}.} 
\end{table} 
\end{center}

\begin{center} 
\begin{table} [H]
\begin{tabular}{c | c  c c  c } 
Country & & & &  \\
\hline
Asia & \includegraphics[width = 0.15\textwidth]{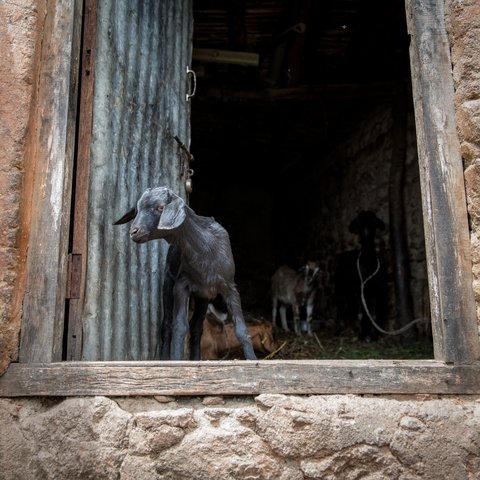} & \includegraphics[width = 0.15\textwidth]{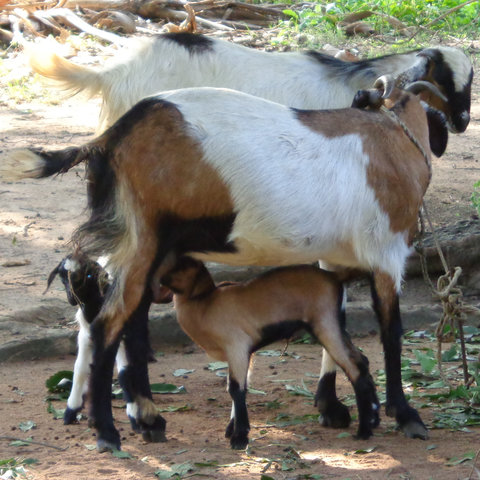} & \includegraphics[width = 0.15\textwidth]{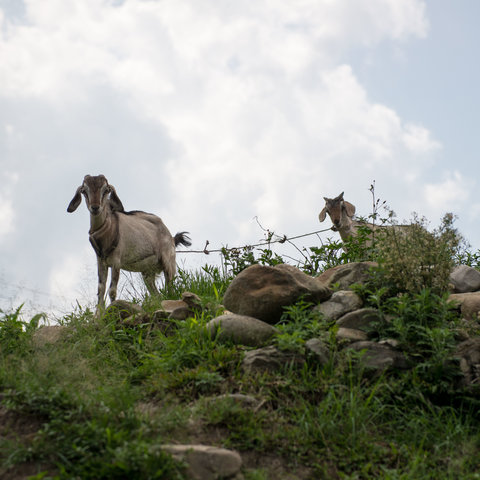} & \includegraphics[width = 0.15\textwidth]{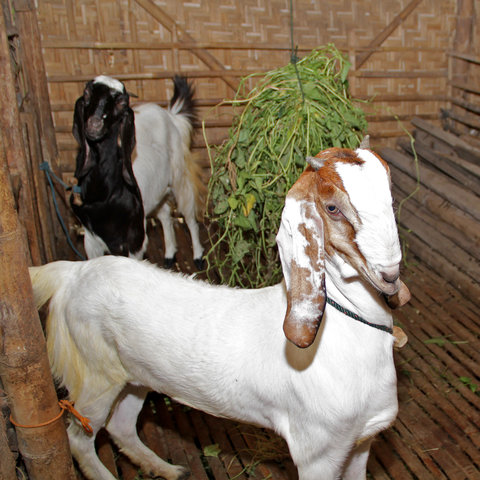} \\ 
 \hline 
 \\Africa & \includegraphics[width = 0.15\textwidth]{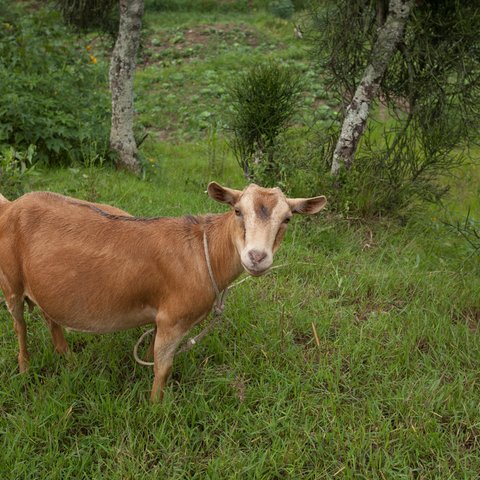} & \includegraphics[width = 0.15\textwidth]{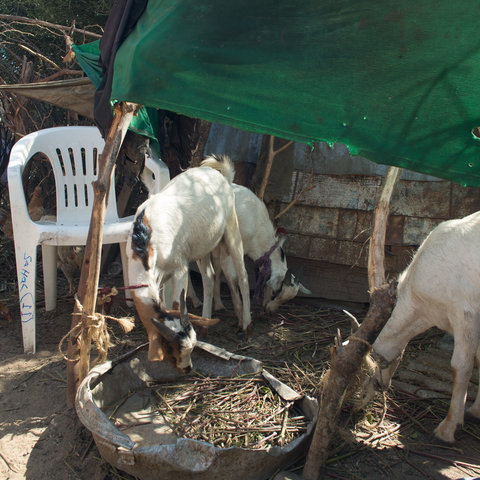} & \includegraphics[width = 0.15\textwidth]{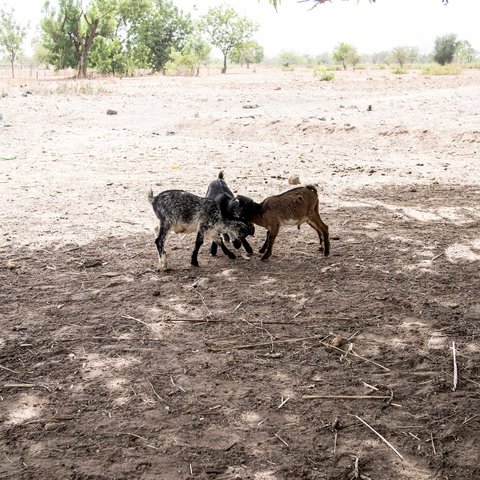} & \includegraphics[width = 0.15\textwidth]{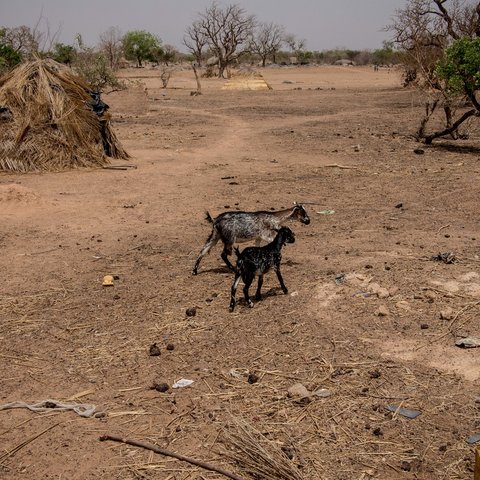} \\ 
 \hline 

\end{tabular} 
\caption{Examples of goat images. Our factors surfaced that images of goats in Dollar Street between regions differed most among \textit{pattern}, \textit{color}, \textit{subcategory}} 
\label{app:goats_jsd}
\end{table} 
\end{center} 

In Figures \ref{app:goats_jsd} and \ref{app:diapers_jsd} we show example images from classes and regions that were found to have some the starkest difference in factors, as measured by JSD.

\end{document}